\documentclass{article}

\usepackage{natbib}
\setcitestyle{numbers,square,citesep={,}}

\usepackage[preprint]{neurips_2024}
\usepackage{multirow}

\usepackage[utf8]{inputenc} %
\usepackage[T1]{fontenc}    %
\usepackage{hyperref}       %
\usepackage{url}            %
\usepackage{booktabs}       %
\usepackage{amsfonts}       %
\usepackage{nicefrac}       %
\usepackage{microtype}      %
\usepackage{xcolor}         %

\usepackage{ulem}
\usepackage{graphicx}
\usepackage{amsmath}
\usepackage{utfsym}
\usepackage{colortbl}

\title{Exploring the Causality of \\
End-to-End Autonomous Driving}

\newcommand{\methodname}{DriveInsight}

\author{
    Jiankun Li \quad Hao Li \quad Jiangjiang Liu \quad Zhikang Zou \quad Xiaoqing Ye\textsuperscript{$\dagger$}   \\
    \textbf{Fan Wang \quad Jizhou Huang\textsuperscript{$\dagger$ $*$} \quad Hua Wu \quad Haifeng Wang} \\
    Baidu Inc., China \\
    \texttt{\{lijiankun02, lihao57, yexiaoqing, huangjizhou01\}@baidu.com}
}

\begin{document}

\maketitle

\renewcommand{\thefootnote}{\fnsymbol{footnote}}
\footnotetext[2]{Corresponding authors.}
\footnotetext[1]{Project lead for end-to-end autonomous driving.}

\begin{abstract}
Deep learning-based models are widely deployed in autonomous driving areas, especially the increasingly noticed end-to-end solutions. However, the black-box property of these models raises concerns about their trustworthiness and safety for autonomous driving, and how to debug the causality has become a pressing concern. Despite some existing research on the explainability of autonomous driving, there is currently no systematic solution to help researchers debug and identify the key factors that lead to the final predicted action of end-to-end autonomous driving. In this work, we propose a comprehensive approach to explore and analyze the causality of end-to-end autonomous driving. First, we validate the essential information that the final planning depends on by using controlled variables and counterfactual interventions for qualitative analysis. Then, we quantitatively assess the factors influencing model decisions by visualizing and statistically analyzing the response of key model inputs. Finally, based on the comprehensive study of the multi-factorial end-to-end autonomous driving system, we have developed a strong baseline and a tool for exploring causality in the close-loop simulator CARLA. It leverages the essential input sources to obtain a well-designed model, resulting in highly competitive capabilities.
As far as we know, our work is the first to unveil the mystery of end-to-end autonomous driving and turn the black box into a white one. Thorough close-loop experiments demonstrate that our method can be applied to end-to-end autonomous driving solutions for causality debugging.
Code will be available at \href{https://github.com/bdvisl/DriveInsight}{https://github.com/bdvisl/DriveInsight}.
\end{abstract}

\section{Introduction}
\label{sec:introduction}

Over the past decade, the field of autonomous driving based on deep neural networks has experienced remarkable growth, encompassing advancements in both academia and industry. Despite the strong representation ability, deep learning suffers from a lack of transparency, making it difficult to identify issues. For the task of autonomous driving, which entails exceptionally high safety requirements, the system's black-box behavior markedly diminishes trust, consequently restricting its practical applications. Therefore, elucidating and addressing causality within such systems is in high demand and remains an unresolved challenge.

Recent advances in interpretability methodologies provide a promising means to comprehend the intricacies of this complex system, garnering increasing interest. Through the integration of natural language descriptors to steer the entirety of decision-making and action processes within autonomous control modules, these methodologies facilitate a more intuitive and comprehensible interpretation of the network's prognostications. Nevertheless, the end-to-end architecture of autonomous driving encompasses multiple modules, with the specific impact of each module on the final decision output remaining unclear. This lack of clarity underscores the need for a systematic analytical framework to help researchers debug and thus hinder the field's progress.

In this work, we propose a comprehensive approach to debug and analyze the causality of end-to-end autonomous driving. The key idea is to assess the individual contributions of each factor and find explanations regarding the most influential features that determine the final predicted action. Similar to the decision-making process of human driving, the final prediction of action / control by end-to-end autonomous driving models is often multi-factorial. For instance, when making an unprotected left turn, the smart agent needs to simultaneously take into account traffic lights, oncoming straight-moving vehicles, pedestrians crossing the road, and static lane lines, etc. Misattribution is common in real-world imitation learning settings. Hence, we conduct a quantitative ablation experiment analysis on the roles of key components and proposed two forms of qualitative analysis methods: counterfactual intervention and response visualization. Finally, based on the comprehensive study of the multi-factorial end-to-end autonomous driving system, we offer a strong baseline and a tool for debugging the causality in close-loop simulator CARLA. It leverages essential input sources to obtain a robust and well-designed model, which not only achieves competitive results but also provides interpretability of the prediction.

As far as we know, our work is the first to unveil the mystery of end-to-end autonomous driving to turn the black box into a white one.
We conduct thorough experiments for driving scenes and show that our method can be applied to end-to-end autonomous driving solutions to diagnose casualty issues effectively.

To sum up, our contributions are as follows:
\begin{itemize}
\item{
We present the first debugging and analyzing solution and baseline to unveil the mystery of black-box end-to-end autonomous driving by explicitly explaining the causality of multi-factor decisions.}
\item{
We conduct detailed quantitative ablation and counterfactual intervention experiments and propose two types of response visualization methods: component-level visualization, which is thoroughly analyzed from the perspectives of temporal consistency of responses and scenario relevance, and activation map visualization, which illustrates the spatial distribution of the impact of semantic features.
}
\item{
By first adopting the counterfactual reasoning to qualitatively figure out the most influential features that lead to the final predicted action, and then applying attention-based strategies to quantitatively analyze the contribution of each factor to adjust the end-to-end model, we are able to get a comprehensive understanding of the decision process.
}

\end{itemize}

\section{Related Works}
\label{sec:related}

\subsection{End-to-End Autonomous Driving}
With the growing computing power, the availability of massive data, closed-loop simulation, and evaluation tools, end-to-end autonomous driving system is getting increasing attention both in the academic and industrial areas. 
Given raw sensor data as input, the end-to-end framework directly predicts the planning and control signals by a single model, instead of traditional cascading multi-module designs. Benefiting from the data-driven paradigm and joint optimization, the cumulative error is eliminated to perform more intelligently like a human does.
We only focus on the recent progress in end-to-end autonomous driving and for the complete review, please refer to \cite{chen2023end,gao2024survey,lan2023end}.

In the industry, the most noticeable end-to-end solution for L2+ autonomous driving must be the recently released Tesla FSD V12, which has occupied increasing driving miles due to Telsa's continuously improving AI training capacity.
Wayve's AV2.0 pioneers a new era of autonomous driving by developing end-to-end AI foundation models, especially the language model LINGO\cite{LINGO-1,LINGO-2} to increase the transparency in end-to-end reasoning and decision-making, as well as the generative world model GAIA-1\cite{hu2023gaia} to generate realistic driving videos from text, action, and video prompts to accelerate training and validation.

In the academic community, the planning-oriented end-to-end pipeline, UniAD\cite{hu2023planning}, turns traditional individually cascaded modules into a joint optimization pipeline to eliminate accumulative errors. However, it is only validated in the open-loop dataset rather than closed-loop evaluation. 
A coarse-to-fine progressive paradigm is proposed in ThinkTwice\cite{jia2023think} to emphasize the importance of the decoder in end-to-end transformer-based pipelines.
VADv2\cite{chen2024vadv2} unifies input sensor data as scene token embeddings to predict the probabilistic distribution of action, which achieves state-of-the-art closed-loop performance on the CARLA benchmark.

\subsection{Causality of End-to-End Autonomous Driving}

A mass of work investigates the interpretability, explainability, or causal inference for deep learning models since it plays an essential role in debugging and providing insights into a model's decision. We mainly focus on the more complex ``black-box'' end-to-end autonomous driving frameworks and introduce several categories of interpretability/causality. 
(1) Leveraging the power of large language models to provide high-level explanations by either formulating the driving task as a visual question answering (VQA) problem\cite{sima2023drivelm,LINGO-1,xu2023drivegpt4,atakishiyev2023explaining}, or outputs explanations of decision-making \cite{wang2023drivemlm,LINGO-2,wang2023empowering}. However, there is still space for improvement of linguistic-aided models for driving systems due to fictitious hallucinations and inaccurate explanations.
(2) Build causal attention/saliency models to indicate which input factors or parts have high influences on the final predictions\cite{Kim_Canny_2017,araluce2024leveraging,chitta2021neat}. For example, \cite{feng2024polarpoint} interprets self-driving decisions and improves safety by paying more attention to the regions that are near the ego vehicle. Attention or saliency approaches provide qualitative clues about the model's focus but lacking of quantitative effects makes the interpretability limited.
(3) Counterfactual explanation intends to figure out the distinguished input features for decision-making, by envisioning modification of those features would cause the model to make a different prediction\cite{zemni2023octet,jacob2022steex,li2020make}.
(4) Interpretable auxiliary tasks, in addition to the final driving policy task, are often constructed to provide interpretable information for analysis and help the model gain a high-level understanding of the scene. The auxiliary tasks usually decode the latent feature representation into perception predictions such as object detection\cite{hu2023planning,jaeger2023hidden,chen2022learning}, semantic segmentation\cite{chitta2022transfuser,kim2021toward}, depth estimation\cite{chitta2022transfuser,chitta2021neat}, motion prediction\cite{chen2022learning,wang2021learning}. 
(5) Causal identification. Given multiple temporal inputs such as the ego car's past states, routing, and sensor data, end-to-end models can easily suffer from causal confusion or the copycat problem in complex scenarios, which means the model cannot distinguish spurious correlations from true factors.
To alleviate the causal confusion problem, the pioneer work\cite{de2019causal} 
learns a mapping from causal graphs to policies, and then uses targeted interventions to efficiently search for the correct policy. PlanTF\cite{cheng2023rethinking} adopts an attention-based state dropout encoder and data augmentation technique to mitigate compounding errors. ChauffeurNet\cite{bansal2018chauffeurnet} solves the causal confusion issue by using the past ego-motion as intermediate bird’s-eye-view (BEV) abstractions and randomly dropping it out during training. \cite{chuang2022resolving} solves the copycat issue by a copycat-free memory extraction module via residual action prediction.

However, most of the above-mentioned works are not specially tailored for end-to-end autonomous driving frameworks and can only improve the traceability of end-to-end's ``black box'' attribute to a certain degree, instead of correctly identifying the most related causal factors to design a real causal model. In consequence, our main motivation is to present debugging and analyzing tools for explicitly explaining the causality of multi-factor complex end-to-end applications. This work first qualitatively identifies the causal factors, then quantitatively analyzes the contribution of each factor to obtain a robust and comprehensive control of end-to-end autonomous driving.

\section{Method}
\label{sec:method}

\begin{figure}[t]
	\begin{center}
		\includegraphics[width=0.95\linewidth]{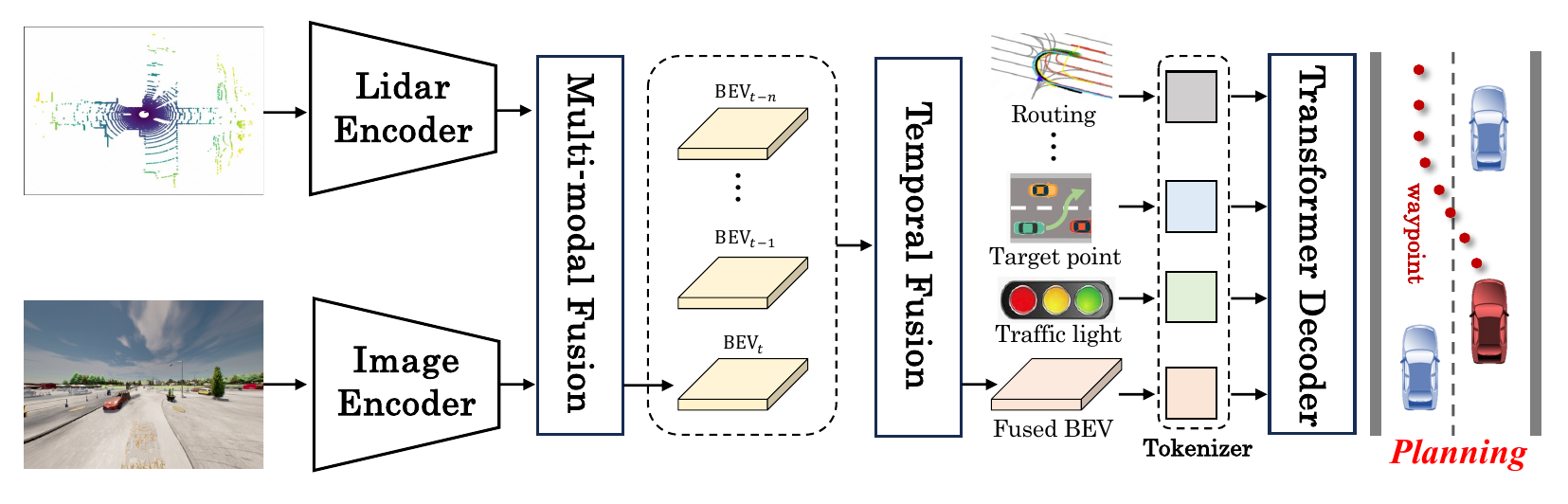}
	\end{center}
    \caption{Overall architecture of our {\methodname} framework. The LiDAR point cloud and multi-view images are processed separately in their respective encoders. Then the resultant features are then fed into the multi-modal fusion and temporal fusion modules sequentially to get the fused BEV features. Besides, we transform other sensor information, including traffic signs, target points, command, routing, and \textit{etc.} into environmental token embeddings. Along with BEV tokens, we sent all these tokens into transformer decoder to predict future trajectories.}
	\label{fig:overview}
\end{figure}

In this section, we delineate the proposed framework for the end-to-end autonomous driving model, named {\methodname}, whose overall architecture is illustrated in Figure~\ref{fig:overview}. Given multi-view images and point cloud input, we first introduce modality-specific encoders to individually extract and transform their distinctive features into BEV representations (Sec. \ref{sec:M-SE}). Subsequently, leveraging multi-modal and temporal fusion modules, we amalgamate these representations to derive unified BEV features (Sec. \ref{sec:F-E}). Lastly, a planning decoder is applied to forecast future trajectories of the ego agent based on the generated BEV tokens and other environmental indicators (Sec. \ref{sec:P-D}).

\subsection{Modality-Specific Encoders}
\label{sec:M-SE}

\noindent\textbf{Camera Encoder.} For the multi-view camera images, we first employ an image backbone architecture (such as ResNet~\cite{he2016deep}) with Feature Pyramid Network, to extract multi-scale image features rich in semantics. Following the widely adopted LSS \cite{philion2020lift}, we use the estimated depth to lift multi-view features to 3D frustums and splat frustums onto a reference plane to generate the BEV features. Specifically, the procedure begins with depth prediction network (DepthNet) predicting the discrete depth distribution for each pixel, which is then used to scatter each pixel into discrete points along the camera ray. At each point, the resultant feature is determined as the outcome of the predicted depth multiplied by the corresponding pixel feature. Within each grid in the BEV feature aggregation is performed using Frustum Pooling, which incorporates features from the points located within the grid. 

\noindent\textbf{Lidar Encoder.} For a given LiDAR point cloud, we first voxelize the input points into uniform bins and extract local 3D shape information in the voxel space using a series of 3D sparse convolutional blocks, consistent with established methodologies in the field. Next, we adopt an hourglass convolutional network as the BEV feature extractor, flattening the 3D features into a 2D BEV view to capture BEV representations rich in contextual information. To maximize the utilization of multi-scale semantics, we employ a feature pyramid network to integrate features from various hierarchical levels, thereby producing scale-aware BEV output features.
\subsection{Fusion Encoders}
\label{sec:F-E}
\noindent\textbf{Multi-Modal Fusion.} Upon the conversion of all sensory features into a unified BEV representation, we employ multi-modal fusion techniques to amalgamate two distinct sets of features, yielding the fused multi-modal features. Initially, a sequence of 2D convolutional layers is employed to standardize two distinct BEV features respectively to a uniform dimension, which are subsequently concatenated and processed through a succession of 2D convolutional layers. To enhance channel-wise interaction, multiple Squeeze-and-Excitation (SE) blocks~\cite{hu2018squeeze} are applied to manipulate the fused features. Given that the solitary direct supervision in the final planning prediction is insufficient to effectively address the intricacies of the high-dimensional multi-sensor input, we introduce supplementary feature-level supervision for the BEV feature map inspired by DriveAdapter~\cite{jia2023driveadapter}. 

\noindent\textbf{Temporal Fusion.} In order to fully leverage the extensive historical context, we have developed temporal fusion modules that align and integrate temporal cues to achieve more accurate predictions. To start with, we construct a memory bank $Q$ to store contextual features extracted from adjacent frames and relative pose. Note that the features corresponding to each frame within the memory bank are mapped to the coordinate system of the present frame through pose transformation. Upon acquiring BEV features of the current frame, we concatenate these features with all features stored in the memory bank and apply a convolutional layer to reduce the channel dimension for saving computation resources. Subsequently, an SE block is utilized to promote interaction, thereby facilitating the derivation of temporally fused features for the current frame. These fused features are subsequently incorporated into the memory bank, while the earliest frame is removed to effectuate the necessary update of the bank.

\subsection{Planning Decoders}
\label{sec:P-D}
The planning decoder receives two components as input: the first is BEV features, which succinctly model the perception of the current environment; the second component comprises additional structured information, which predominantly includes three categories: ego vehicle status, environmental information, and navigation information. Ego vehicle status information includes the speed at the current moment and historical moments, while environmental information includes structured information about high-definition maps, obstacles, traffic lights, and stop signs. Typically, such information can be predicted by the model through auxiliary task modules. However, to streamline the task and concentrate on casualty analysis itself, the relevant data is directly provided by the simulation environment in this study. Navigation information includes command, target point, and routing. The command represents information from the high-level planner, such as STRAIGHT, RIGHT, LEFT, etc. The target point indicates the position and orientation of the target, while routing is a collection of dense navigation points at the lane level. BEV features and structured information are separately encoded through MLPs and then concatenated to obtain the final enhanced feature, which provides rich and necessary guidance for downstream behavior planning. To avoid the shortcut learning problem, we employ the dropout strategy during the training phase, which involves randomly masking certain inputs with a certain probability. During the testing phase, the dropout rate is set to 0. Following UniAD~\cite{hu2023_uniad}, we use a query-based design, using an ego query to perform cross-attention with the above features, ultimately obtaining the future trajectory $T$ of the ego vehicle.

\section{Experiments}
\label{sec:exp}

\subsection{Experiment Setup}
To effectively and intuitively evaluate the 
multi-factorial originated end-to-end autonomous driving framework, 
we establish a baseline model called {\methodname} and adopt the popular CARLA~\cite{dosovitskiy17carla} simulator 
with version 0.9.10.1 for training and testing.

\noindent{\textbf{Data Collection and Filtration.}}
We roll out a rule-based expert agent utilizing  
privileged information from the CARLA simulator 
to collect training data at 2~Hz
on 8 towns (\emph{i.e.}, [01, 02, 03, 05, 04, 06, 07, 10]) 
and 14 kinds of weather.
The data collection process generally 
follows the one used in \cite{shao2023safety}. %
For different towns and weathers, the routes are randomly generated 
and the dynamic objects and adversarial scenarios are randomly spawned 
for better data diversity.
For the sensors, we use four RGB cameras (front, left, right, back) 
of the same resolution ($1600 \times 900$) and FOV ($150^{\circ}$), 
one LiDAR with default configuration 
(rotation frequency: 10~Hz, 
upper/lower FOVs: $10^{\circ}/-30^{\circ}$, 
\#channels: 32), IMU (20~Hz), GPS (100~Hz) and speedometer (20~Hz). 
Despite the perception and ego state information, 
the vectorized map, states of the traffic elements,
sparse target point and high-level command, 
and dense routing information are also collected.
The target points are provided as GPS coordinates
following the standard protocol of CARLA, which are
sparse and can be hundreds of meters apart.
The routing is lane-level and can be viewed as 
the interpolated version of the target points
with the resolution of one meter.

We collect a dataset of 3.1M frames (19,105 routes) with all 8 towns to match 
the data amount of existing work \cite{shao2023safety}(3M, 8 towns, 2~Hz)
and \cite{hu2022model}(2.9M, 4 towns, 25~Hz). 
For all routes, as done in \cite{jia2023think} %
we truncate the last few frames 
where the vehicle stops in case of timeout (0.5M frames removed).
Since the expert agent we use is not perfect, 
we additionally filter out the entire route if any infraction happened 
(\emph{i.e.}, agent ran a stop or a red light, collided against object,
got blocked, went outside its route lanes, or deviated from the route).
This post-process removes 1,229 routes.
The final data amount for training is 1.8M frames (17,876 routes).

\noindent{\textbf{Training Strategy.}} We train the model on 32 A800 GPUs for 4 epoches, using a initial learning rate of $1 \times 10^{-4}$ that decayed with cosine annealing policy. The batch size is set to 8 per GPU to fit the graphics memory. We use a simple L1 loss to supervise the model's output waypoints. The point cloud range we use is [-8.0m, -19.2m, -4.0m, 30.4m, 19.2m, 10.0m] following \cite{jia2023think}.

\noindent{\textbf{Benchmark.}}
We assess the efficacy of our approach utilizing the Town05 Long and Town05 Short benchmarks.
The Town05 Long benchmark comprises 10 diverse routes, 
spanning approximately 1 kilometer each, 
presenting a rigorous examination of the model's 
comprehensive adaptability across varied scenarios. 
Conversely, the Town05 Short benchmark features 32 succinct routes, 
with each route spanning 70 meters, 
designed to scrutinize the model's performance in targeted scenarios, 
such as intricate maneuvers like lane changes preceding intersections. 
These benchmarks encapsulate a spectrum of challenges, 
including dynamic agent interactions and adversarial occurrences, 
echoing real-world driving conditions. 
By adhering to these benchmarks, 
we ensure a thorough evaluation of our method's efficacy 
in navigating predefined routes, mitigating collisions, 
and adhering to traffic regulations amidst challenging environments.

\noindent{\textbf{Evaluation Metrics.}}
Following common practice, we utilize three official metrics 
introduced by the CARLA Leaderboard
\footnote{\url{https://leaderboard.carla.org/\#evaluation-and-metrics}}
for evaluation: Route Completion (RC), Infration Score (IS), and 
Driving Score (DS).
Route Completion (RC) represents the percentage of the route distance completed by an agent. 
It accounts for the agent's adherence to the designated route, 
penalizing deviations from the prescribed path.
Infraction Score (IS) quantifies infractions committed by the agent, 
including collisions with pedestrians, vehicles, road layouts, 
and violations of traffic signals. 
Each infraction incurs a penalty coefficient, 
reducing the overall score proportionally to the severity of the infraction.
Driving Score (DS) serves as a composite metric, 
capturing both driving progress and safety. 
It is calculated as the product of Route Completion and Infraction Score, 
providing a comprehensive assessment of the agent's performance.
By employing these metrics, we aim to provide a holistic evaluation of 
the driving behavior exhibited by each agent, 
fostering a deeper understanding of their performance in diverse scenarios.

\subsection{Ablation Study and Analysis}
\subsubsection{Effectiveness of Prompts of Planning Decoder}
In this part, we conduct extensive experiments to demonstrate the effect of key components in the planning decoder. By systematically setting each component to zero individually, we analyze its specific contribution to the final performance. The results are shown in Table \ref{table:main_abstudy}. We can see that the exclusion of BEV features results in significant performance degradation from the table. This is because a fundamental aspect of autonomous driving is the scene modeling of the surrounding environment. The primary advantage of BEV technology lies in its capability to provide an intuitive representation of the scene distribution, thereby exerting the most significant influence on the accuracy and efficacy of the final trajectory planning. In contrast, the final performance does not show any significant change in the absence of map, command, stop signs, or obstacles, which underscores the insignificance of these variables in determining the final planning results. 

For the navigation relevant components including routing and target points, the absence of them seriously drop the final performance especially the route completion. To elucidate further, the driving scores exhibit a notable degradation, decreasing from 95.30 to 20.64 and 32.64, respectively. This reveals the paramount importance of global target-orientated guidance in the modeling of autonomous driving networks. Furthermore, we compared the impact of current speed and historical speed information on the model's performance. We found that the contribution of historical speed is quite limited, whereas the current speed information is crucial for the model's motion planning.

\begin{table}[h]
\renewcommand{\arraystretch}{1.2}
\caption{Ablation Study of each component, where Cmd, T.P., C.S., P.S., Obs, S.S, ,T.L. mean 
command, target point, current speed, previous speed, obstacles, stop signs and traffic lights respectively.
}\label{table:main_abstudy}
\centering
\resizebox{\textwidth}{!}{
\begin{tabular}{c|cc|ccc|cc|ccc|ccc}
\hline
\multirow{2}{*}{Methods} & \multicolumn{10}{c|}{Key Components} & \multicolumn{3}{c}{Performance} \\ \cline{2-14} 
&   BEV & Map & Routing & Cmd & T.P. & C.S. & P.S. & Obs & S.S. & T.L. & DS $\uparrow$ & RC$\uparrow$  & IS$\uparrow$  \\ \hline
\rowcolor{gray!10} Baseline~~         & & & & & & & & & & &  95.30 & 99.26 & 0.96  \\ 
\hline
(a)~~ & \usym{2715} & & & & & & & & & & 51.66 & 80.82 & 0.66 \\
(b)~~ & & \usym{2715} & & & & & & & & & 94.55 & 99.78 & 0.94 \\ 
(c)~~ & & & \usym{2715} & & & & & & & & 20.64 & 29.03 & 0.64 \\ \hline
(d)~~ & & & & \usym{2715} & & & & & & & 96.61 & 99.11 & 0.97 \\
(e)~~ & & & & & \usym{2715} & & & & & & 32.64 & 47.47 & 0.69 \\ 
(f)~~ & & & & & & \usym{2715} & & & & & 88.61 & 93.36 & 0.93 \\ \hline
(g)~~ & & & & & & & \usym{2715} & & & & 93.92 & 99.59 & 0.94 \\ 
(h)~~ & & & & & & & & \usym{2715} & & & 94.86 & 98.43 & 0.95 \\
(i)~~ & & & & & & & & &  \usym{2715} & & 94.91 & 99.25 & 0.95 \\
(j)~~ & & & & & & & & & &  \usym{2715} & 49.74 & 81.60 & 0.68 \\
\hline
\end{tabular}
}
\end{table}

\newcommand{\addFig}[1]{\includegraphics[width=0.245\linewidth]{figs/prompt_edit/#1}}
\newcommand{\addFigCrop}[1]{\includegraphics[trim=350 0 350 0, clip, width=0.245\linewidth]{figs/prompt_edit/#1}}
\newcommand{\addFigs}[1]{\addFigCrop{rgb_front/#1.png} & \addFig{topdown/#1.png} & \addFig{birdview/#1.png} & \addFig{vis/#1.png}}
\begin{figure}
  \centering
  \footnotesize
  \setlength\tabcolsep{0.3mm}
  \renewcommand\arraystretch{1.0}
  \begin{tabular}{ccccc}
    \rotatebox{90}{~~~~~~~~~~~~$t_0=10$} & 
    \addFigs{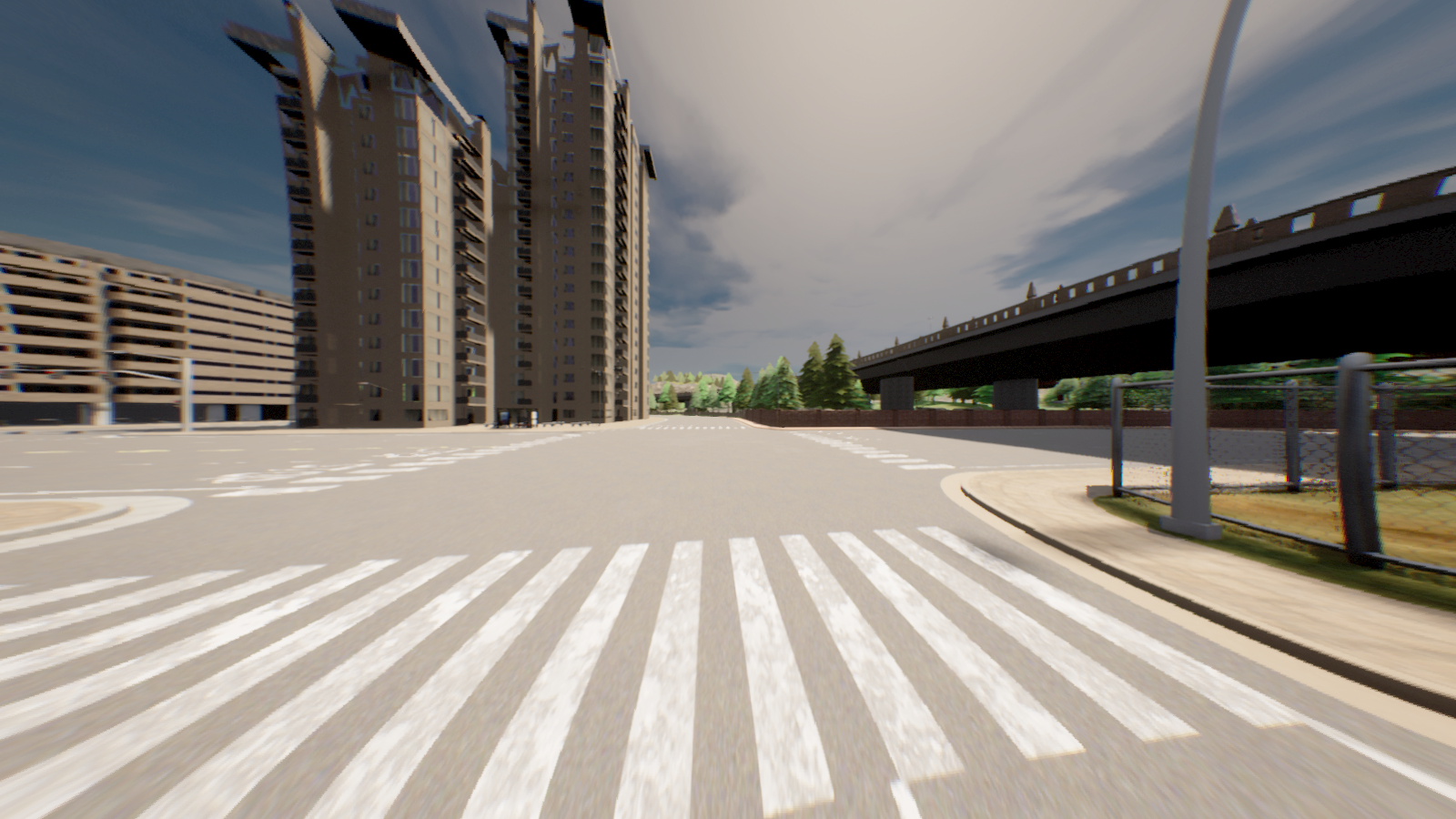} \\
    \rotatebox{90}{~~~~~~~~~~~~$t_1=62$} & 
    \addFigs{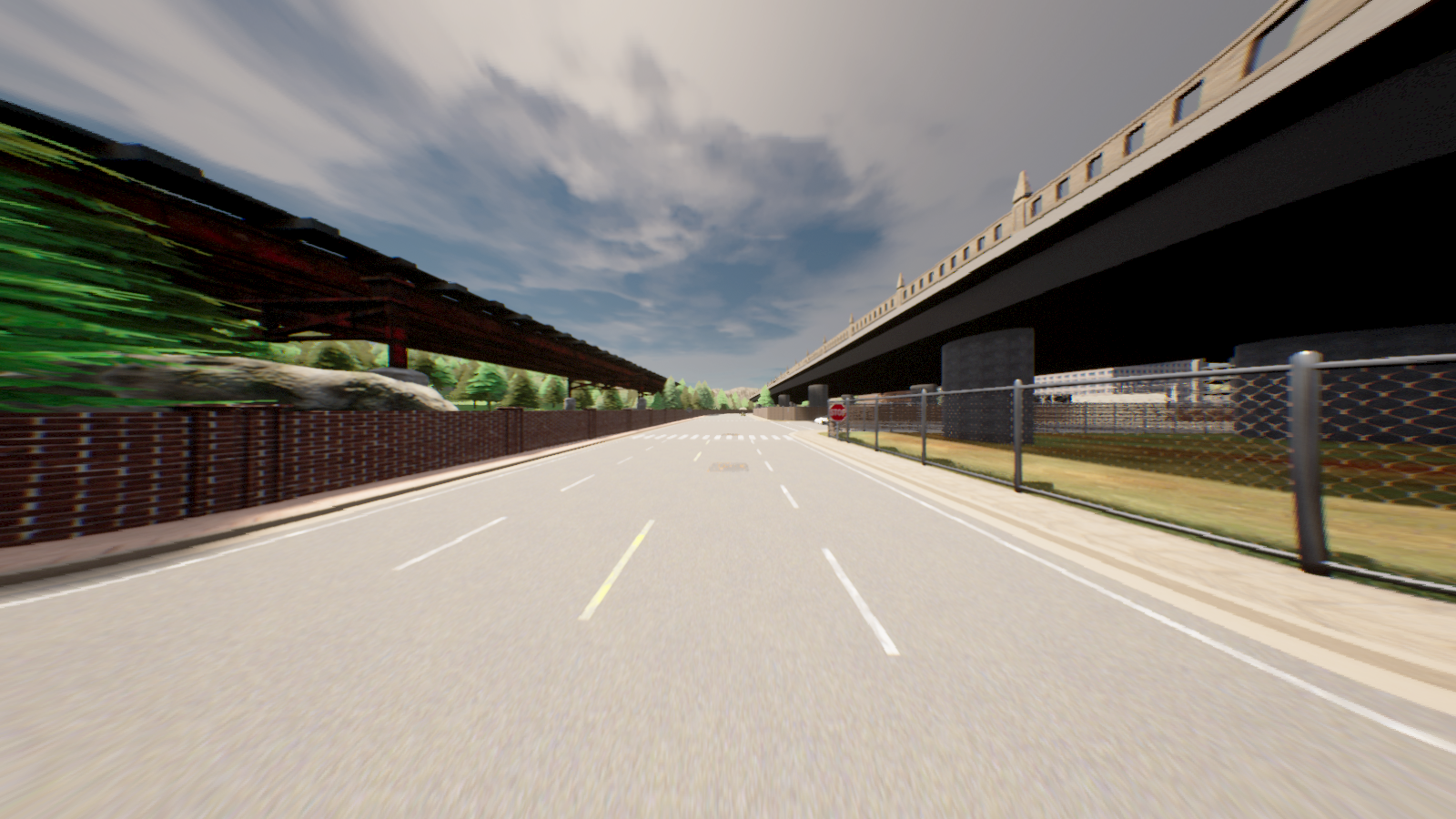}  \\
    \rotatebox{90}{~~~~~~~~~~~~$t_2=102$} & 
    \addFigs{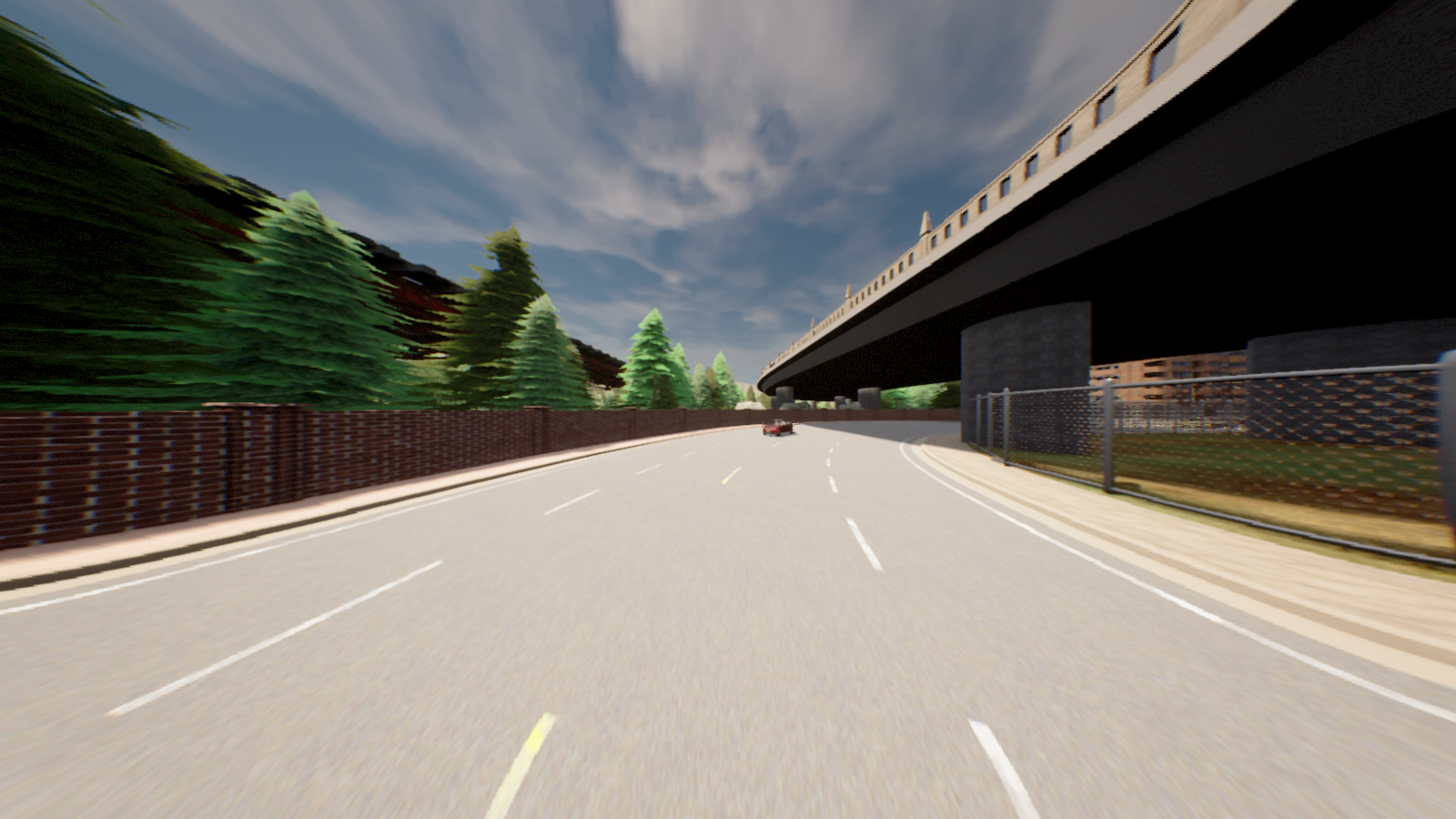} \\
    & Front RGB Camera & RGB Topdown & Structured Visualization & Structured Information \\
  \end{tabular}
  \vspace{3pt}
  \caption{Visualizations of different simulation time steps.The last column shows the visualization of the point cloud and component information. The green curves represent routing, the red dots indicate the target point, the dark blue lines represent the vectorized map, and the light blue rectangles indicate obstacles. Simulation time steps $t_0$, $t_1$, and $t_2$ correspond to the three sampling moments in Figure~\ref{fig:prompt_edit_grad} and Figure~\ref{fig:prompt_edit_weight} (represented as blue, orange, and green in the figures), respectively.}
  \label{fig:prompt_edit_pixel}
\end{figure}

\renewcommand{\addFigs}[1]{\includegraphics[width=1\linewidth]{figs/prompt_edit/#1/#1.pdf}}
\begin{figure}
  \centering
  \footnotesize
  \setlength\tabcolsep{0.3mm}
  \renewcommand\arraystretch{0.9}
  \begin{tabular}{c}
    \addFigs{grad}
  \end{tabular}
  \caption{Visualizations of the gradients \emph{w.r.t.} different tokens by simulation time steps. The gradients in the x and y directions are represented by $G_x$ and $G_y$, respectively. The horizontal axis represents the time elapsed along the current route. We sampled three representative moments, denoted as $t_1$, $t_2$, and $t_3$, indicated in the graph by blue, orange, and green vertical lines, respectively. }
  \label{fig:prompt_edit_grad}
\end{figure}

\renewcommand{\addFigs}[1]{\includegraphics[width=1\linewidth]{figs/prompt_edit/#1/#1.pdf}}
\begin{figure}
  \centering
  \footnotesize
  \setlength\tabcolsep{0.3mm}
  \renewcommand\arraystretch{0.9}
  \begin{tabular}{c}
    \addFigs{weight}
  \end{tabular}
  \caption{Visualizations of the activation \emph{w.r.t.} different attention heads of different simulation time steps. The three colors in the histogram correspond to the sampling time points in Figure 3. The red line represents the average response value of different components over the observation time interval.}
  \label{fig:prompt_edit_weight}
\end{figure}

\begin{figure}
  \centering
  \footnotesize
  \setlength\tabcolsep{0.3mm}
  \renewcommand\arraystretch{1.0}
  \begin{tabular}{c}
   \includegraphics[width=0.246\linewidth]{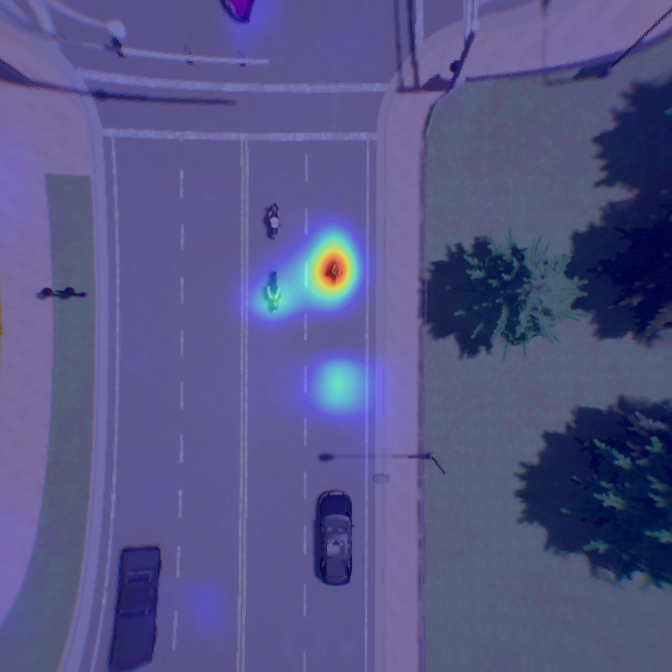}
   \includegraphics[width=0.246\linewidth]{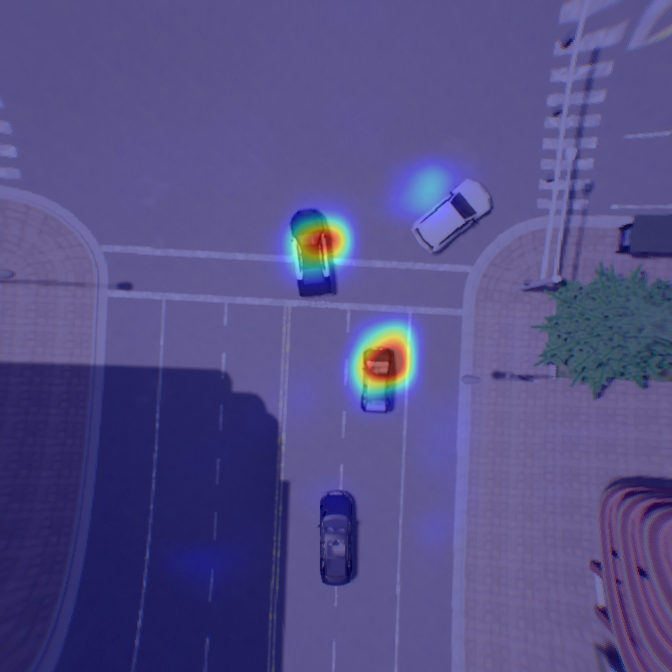}
   \includegraphics[width=0.246\linewidth]{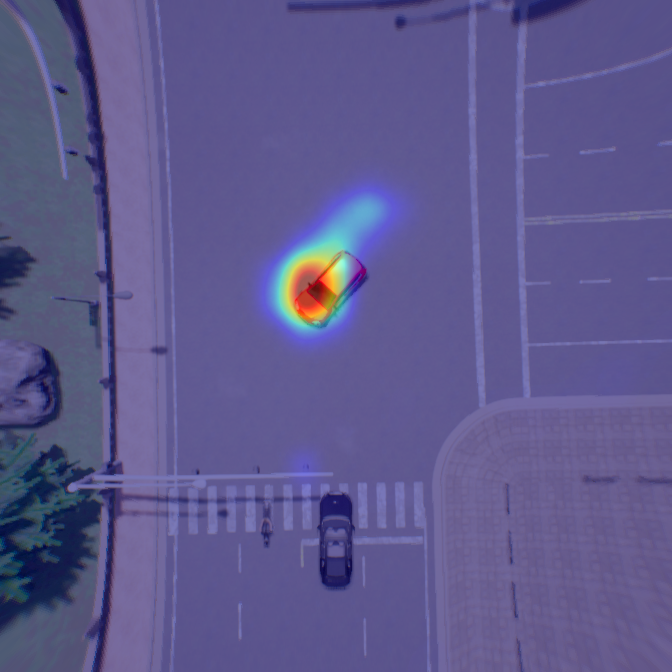}
   \includegraphics[width=0.246\linewidth]{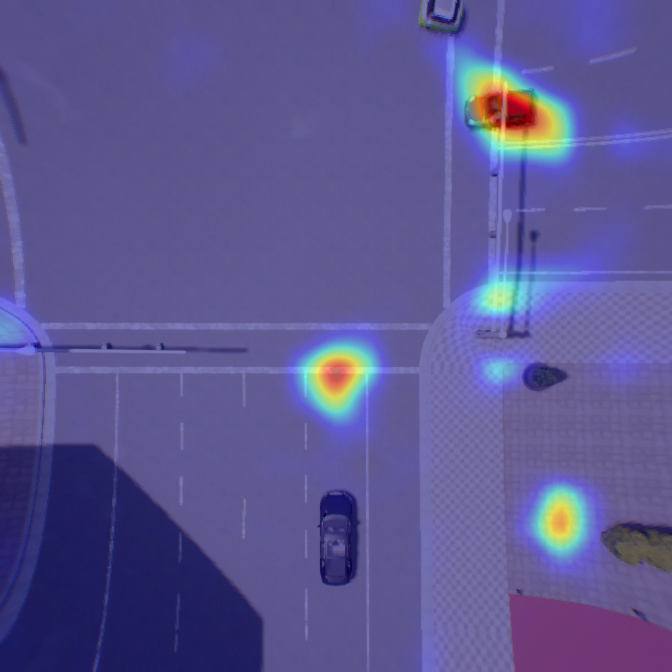} \\
   (a) Activation map of BEV feature.Overlap and aligned with topdown camera in CARLA for visualization. \\
   \includegraphics[width=0.33\linewidth]{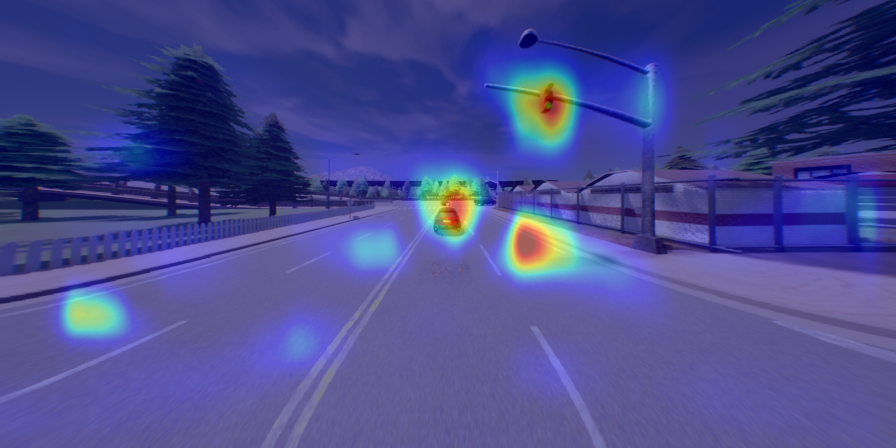}
   \includegraphics[width=0.33\linewidth]{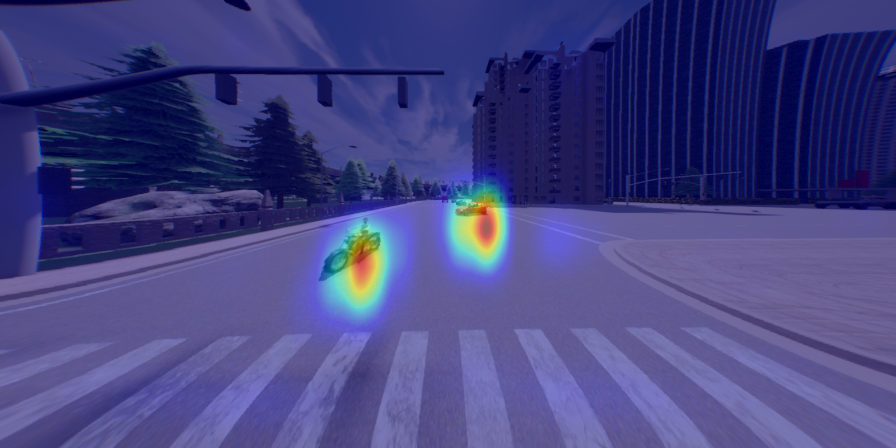}
   \includegraphics[width=0.33\linewidth]{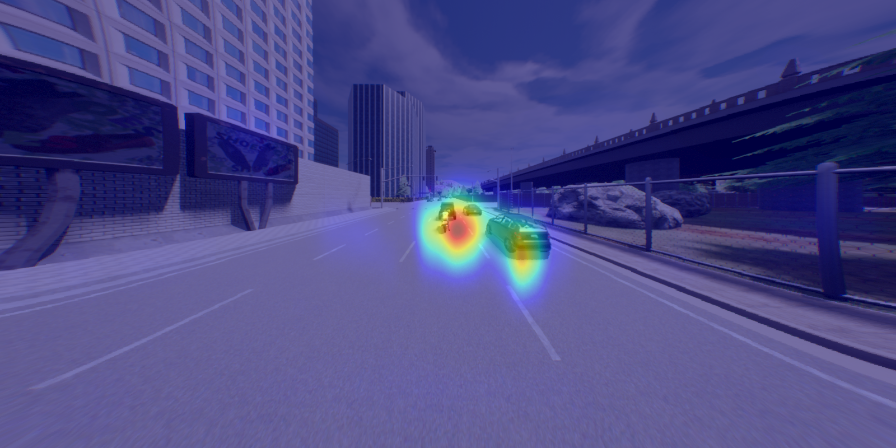} \\
   (b) Activation map of front-view feature. Overlap and aligned with front-view camera in CARLA for visualization. \\   
  \end{tabular}
  \vspace{3pt}
  \caption{Visualization of the activation map}
  \label{fig:activation_map}
\end{figure}

\subsubsection{Prompt Editing}

In this section, we employ counterfactual interventions to manually edit factors that might affect the model's final planning prediction. We construct counterfactual or perturbative prompts targeting the model's input at time $t$ $(O_t, P_t)$, resulting in a new input $(O_t, P_t')$, where $O$ represents observed information and $P$ denotes the prompt information.

By implementing this approach, we observe how the model's prompt behaves under different human interventions, allowing us to qualitatively analyze the actual impact of each component.

\noindent\textbf{Routing and target point.} Only modifying either the routing or the target point alone cannot significantly intervene in the behavior of the ego vehicle. However, when both the routing and target point are simultaneously modified, the behavior of the ego vehicle can be successfully intervened and controlled. This is true even if the modified planned route may not conform to normal driving logic (e.g., driving in the wrong direction or towards the edge of the road). This demonstrates the combined effect of routing and target point on the behavior planning of the ego vehicle.

\noindent\textbf{Current speed.} We modified the current speed to fixed values of 0m/s (stationary) and 10 m/s respectively, and compared these results with the input of the vehicle's actual driving speed. When the given speed is consistently 0m/s, the model tends to predict a set of very close waypoints, which causes the vehicle to mimic the start-up phase, slowly creeping forward. When the given speed is consistently 10 m/s, the predicted waypoints span a significantly larger distance, indicating a noticeable acceleration of the vehicle. Interestingly, even after accelerating, the ego vehicle still performs avoidance maneuvers in response to obstacles.

\noindent\textbf{Map.} We perturbed the structured map information, including overlaying two-dimensional Gaussian noise and applying random masking. Through experiments, we found that the model exhibits strong robustness to perturbations in the map; there was no significant increase for off-road and block case. This may be due to the model having already acquired sufficient road modeling information from the BEV features, leading to a reduced dependency on map inputs.

\noindent\textbf{Traffic lights.} We experimented with editing the color information in the structured prompt for traffic lights and observed that the driving behavior of the ego car was easily influenced by the traffic light color. For instance, when the green light was changed to red, the ego car decelerated and braked before the stop line.

\noindent\textbf{BEV feature.} During the testing phase, we applied random mask to the BEV features passed into the planning decoder. In comparison to the baseline, we found that in cases where there were missing parts of the BEV, even when relevant structural information was provided, the model was more prone to collisions with obstacles, running red lights, entering the wrong lane, and going off-road. The likelihood of the vehicle being blocked also increased significantly. This indicates that BEV features aggregate important environmental perception information, which is necessary for the vehicle to drive safely.

\subsubsection{Visual Analysis of Component Responses}
 
We further analyzed the role of each input by calculating the gradients of different components relative to the output and visualizing the attention weights of the transformer layers, in combination with specific scenarios and the behavior of the ego vehicle.

By analyzing Figure~\ref{fig:prompt_edit_pixel} and Figure~\ref{fig:prompt_edit_grad}, we can draw conclusions about the component-level correlations in specific scenarios, indicating the model’s sensitivity to changes in different components within a fixed scenario. Specifically, we can observe distinct patterns of each token over time by combining environmental perception information. For example, when the vehicle passes through an intersection, the model is more sensitive to traffic lights and stop signs; when an obstacle appears in front of the vehicle, the model is more sensitive to obstacle information and routing; when the vehicle is turning, the roles of command, routing, and the map are more significant. Additionally, we notice that BEV features exhibit high response values in multiple scenarios, such as turning and obstacle avoidance. 

From Figure~\ref{fig:prompt_edit_weight}, we can observe the head-level response of the transformer decoder to different components. The arrangement of the bar charts shows that different heads may correspond to multiple components, and the response preferences for these components exhibit high consistency. For example, $head_3$ might focus more on obstacles and BEV features, while $head_4$ might be more focused on speed. Furthermore, by comparing the bar charts and the curve, we can see that even though a few bar charts and the curve might not align perfectly, the head’s response to components overall maintains temporal consistency within the observation period. Utilizing this consistency, we can more intuitively analyze the model’s black-box behavior and provide reasonable input-related explanations for the model’s responses.

\subsubsection{Visual Analysis of Activation Map}

To further analyze the impact of semantic features as well as its spatial distribution in the intermediate layers on the final prediction,  referring to \cite{selvaraju2017grad}, we calculate the gradient at each position of each channel $k$ in the target feature map with respect to the predicted waypoints in two directions, and perform global average pooling to obtain the weight coefficient for each channel $\alpha_k^c$:

\begin{equation}
\alpha_k^c = \frac{1}{Z} \sum_i \sum_j \frac{\partial p^c}{\partial A_{ij}^k}, \quad c \in \{x, y\},
\end{equation}

where $p$ is the predicted waypoints, $A_{ij}^k$ is channel $k$ of the target feature map and $Z$ is its spatial size. We then use these weight coefficients to compute a weighted average of the target feature map across all channels, followed by calculating the L2 norm in the $x$ and $y$ directions of the waypoints to obtain the final gradient-weighted activation map $F$:

\begin{equation}
F = \left[ \left( \sum_k \alpha_k^x A^k \right)^2 + \left( \sum_k \alpha_k^y A^k \right)^2 \right]^{\frac{1}{2}}.
\end{equation}

We conducted analysis on the activation responses in two types of feature maps. Firstly, we analyzed BEV features, as shown in Figure~\ref{fig:activation_map}(a). We found that the model has a strong perception ability towards obstacles in the driving direction ahead. When the vehicle is waiting to turn at an intersection, the model shows significant responses to the position of the stop line and the traffic flow at the intersection. Additionally, as illustrated in Figure~\ref{fig:activation_map}(b), the actication map of the front-view camera features shows that the model also pays high attention to traffic lights, drivable lane areas, and vehicles at intersections. More examples can be found in the appendix.

\begin{table}[tp]
\renewcommand{\arraystretch}{1.2}
\renewcommand{\tabcolsep}{8.0pt}
\caption{Closed-loop evaluation on the Town05 Long \& Short benchmarks. Our method achieve a competitive driving score while also achieving the highest route completion. The sign \textsuperscript{*} denotes that we  exclude data from town05 in the training set. }
\label{table:closed_loop}
\centering
\resizebox{\textwidth}{!}{
\fontsize{5}{6}\selectfont %
\begin{tabular}{l|ccc|cc|cc}
\hline
\multirow{2}{*}{Method} & \multirow{2}{*}{Modality} & \multirow{2}{*}{Reference} & \multirow{2}{*}{Training frames} & \multicolumn{2}{c|}{Town05 Long} & \multicolumn{2}{c}{Town05 Short} \\ \cline{5-8}
 & & & & DS $\uparrow$ & RC $\uparrow$ & DS $\uparrow$ & RC $\uparrow$ \\
\hline
LBC~\cite{chen2020lbc} & C & CoRL 20 & 150K & 12.3 & 31.9 & 31.0 & 55.0 \\
Transfuser~\cite{chitta2022transfuser} & C+L & TPAMI 22 & 150K & 31.0 & 47.5 & 54.5 & 78.4 \\
ST-P3~\cite{hu2022stp3} & C & ECCV 22 & 150K & 11.5 & 83.2 & 55.1 & 86.7 \\
VAD~\cite{vad} & C & ICCV 23 & 3.0M & 30.3 & 75.2 & 64.3 & 87.3 \\
ThinkTwice~\cite{jia2023think} & C+L & CVPR 23 & 2.2M & 70.9 & 95.5 & - & - \\
MILE~\cite{hu2022model} & C & NeurIPS 22 & 2.9M & 61.1 & 97.4 & - & - \\
Interfuser~\cite{shao2023safety} & C & CoRL 22 & 3.0M & 68.3 & 95.0 & 94.9 & 95.2 \\
DriveAdapter~\cite{jia2023driveadapter} & C+L & ICCV 23 & 2.0M & \textbf{71.9} & 97.3 & - & - \\
Ours & C+L & - & 1.8M & 66.6 & \textbf{100.0} & \textbf{95.3} & \textbf{99.2} \\
Ours\textsuperscript{*} & C+L & - & 1.5M & 64.4 & \textbf{100.0} & 93.2 & 95.8 \\
\hline
\end{tabular}
}
\end{table}

\subsection{Comparison with Other Open-Source Methods}

We conduct closed-loop evaluation on Town05 Long and Town05 Short benchmark in CARLA. As shown in Table \ref{table:closed_loop}, compared to other state-of-the-art methods, our model achieve a competitive driving score while also achieving the highest route completion. Note that, compared to other algorithms such as \cite{jia2023think} and \cite{chitta2022transfuser}, which require manually designed rules to post-process the control signals after the PID (Proportional-Integral-Derivative) controller to avoid violations or getting stuck, our end-to-end approach does not incorporate any manual rules; instead, the waypoints generated by our model are directly converted into control signals through the PID controller.

\section{Conclusion}
\label{sec:conclusion}
In this paper, we introduce a pioneering debugging and analysis solution, designed to demystify black-box end-to-end autonomous driving by explicitly elucidating the causality of multi-factor decisions. Our analysis system is divided into three steps: quantitative analysis of module drop, case analysis of module editing, and visualization of gradient response values. Extensive experiments are conducted using the popular CARLA to validate the reliability of our analytical system. We believe that this system can serve as a benchmark for end-to-end autonomous driving, thereby enhancing the interpretability and reliability of future designs.

\clearpage

{\small
\normalem
\bibliographystyle{plainnat}
\bibliography{egbib}
}

\clearpage

\section{Appendix}

\renewcommand{\addFig}[1]{\includegraphics[width=0.245\linewidth]{figs/case2/#1}}
\renewcommand{\addFigCrop}[1]{\includegraphics[trim=350 0 350 0, clip, width=0.245\linewidth]{figs/case2/#1}}
\renewcommand{\addFigs}[1]{\addFigCrop{rgb_front/#1.png} & \addFig{topdown/#1.png} & \addFig{birdview/#1.png} & \addFig{vis/#1.png}}
\begin{figure}[h]
  \centering
  \footnotesize
  \setlength\tabcolsep{0.3mm}
  \renewcommand\arraystretch{1.0}
  \begin{tabular}{ccccc}
    \rotatebox{90}{~~~~~~~~~~~~$t_0=67$} & 
    \addFigs{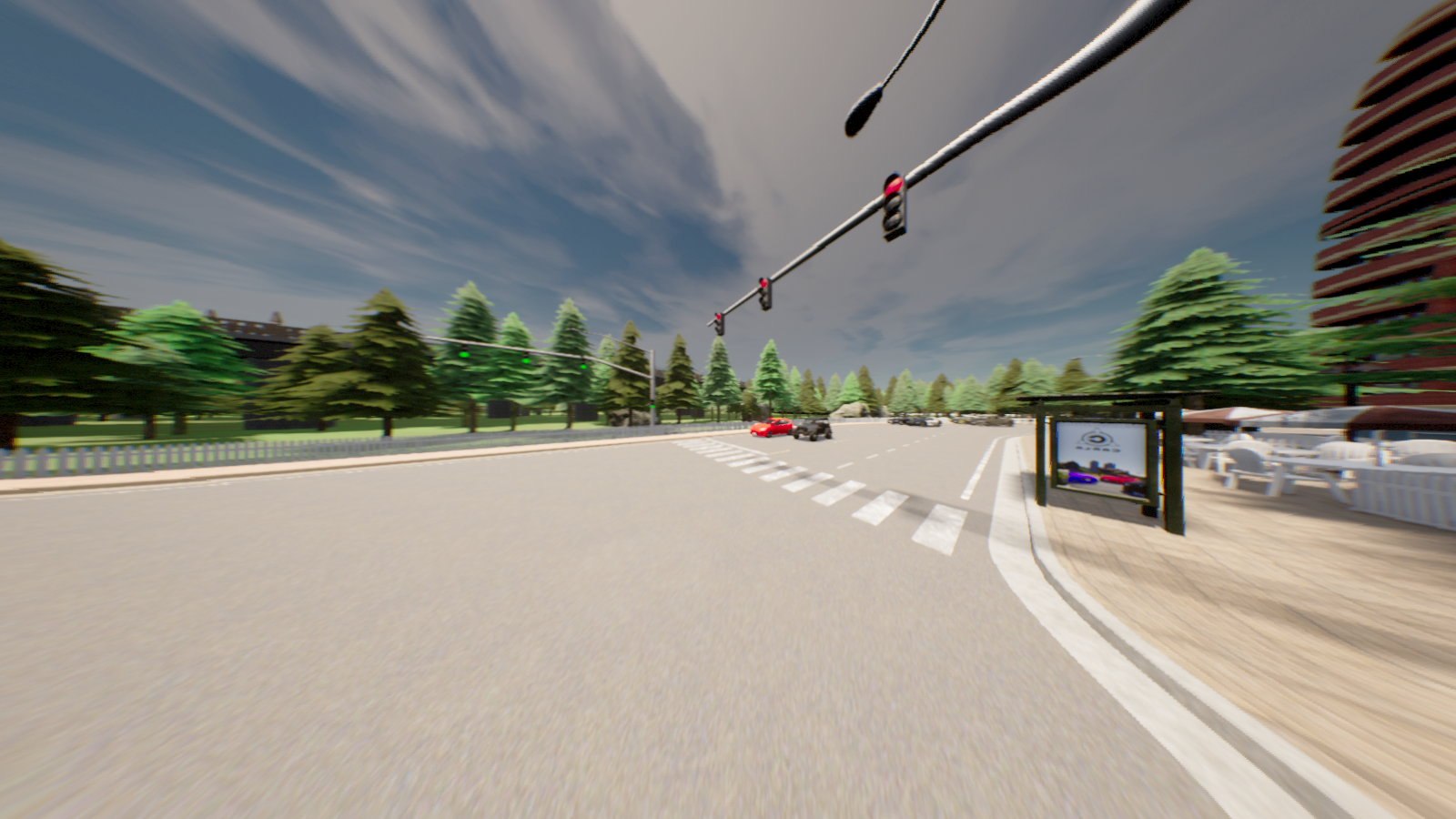} \\
    \rotatebox{90}{~~~~~~~~~~~~$t_1=85$} & 
    \addFigs{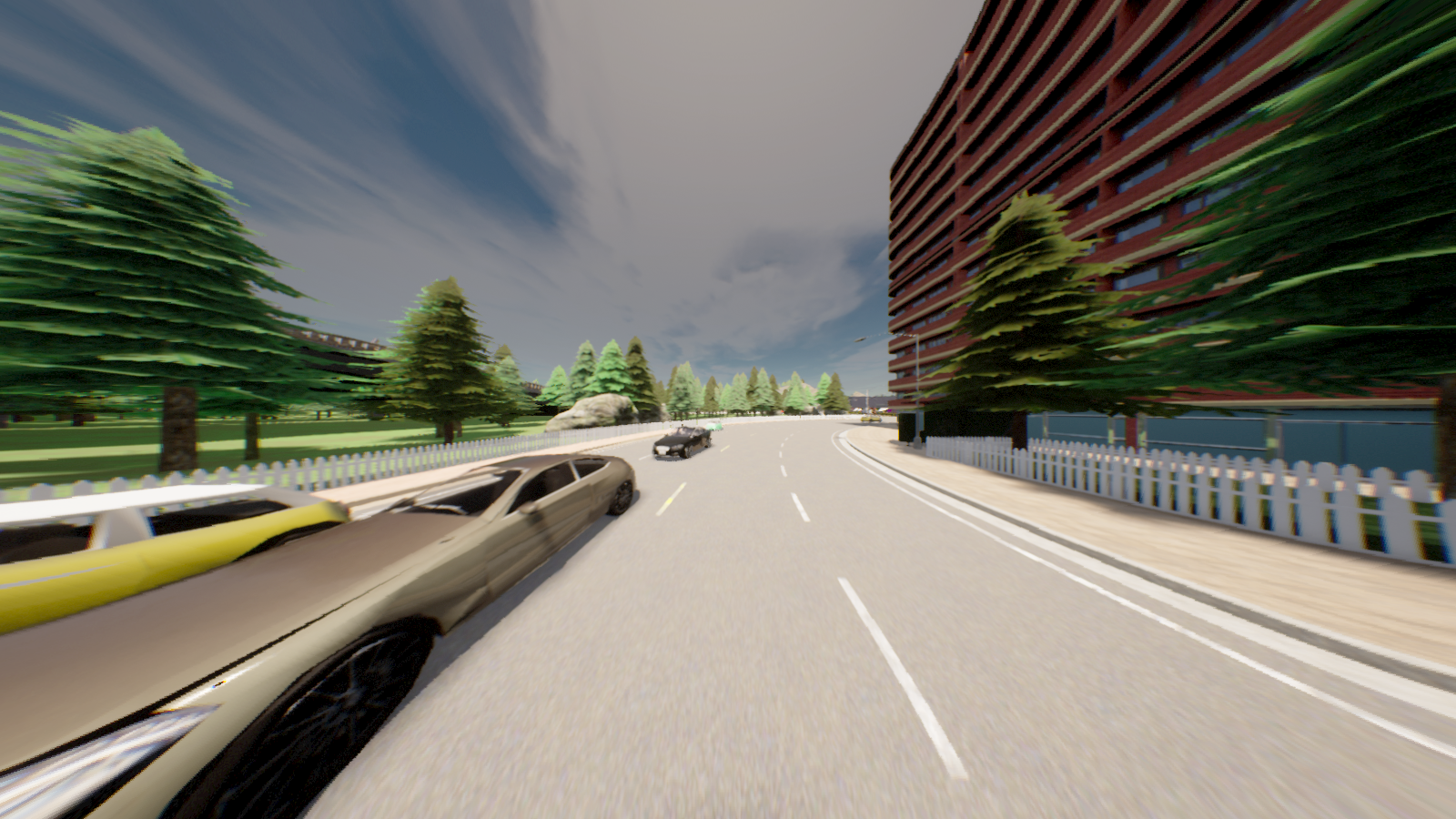}  \\
    \rotatebox{90}{~~~~~~~~~~~~$t_2=127$} & 
    \addFigs{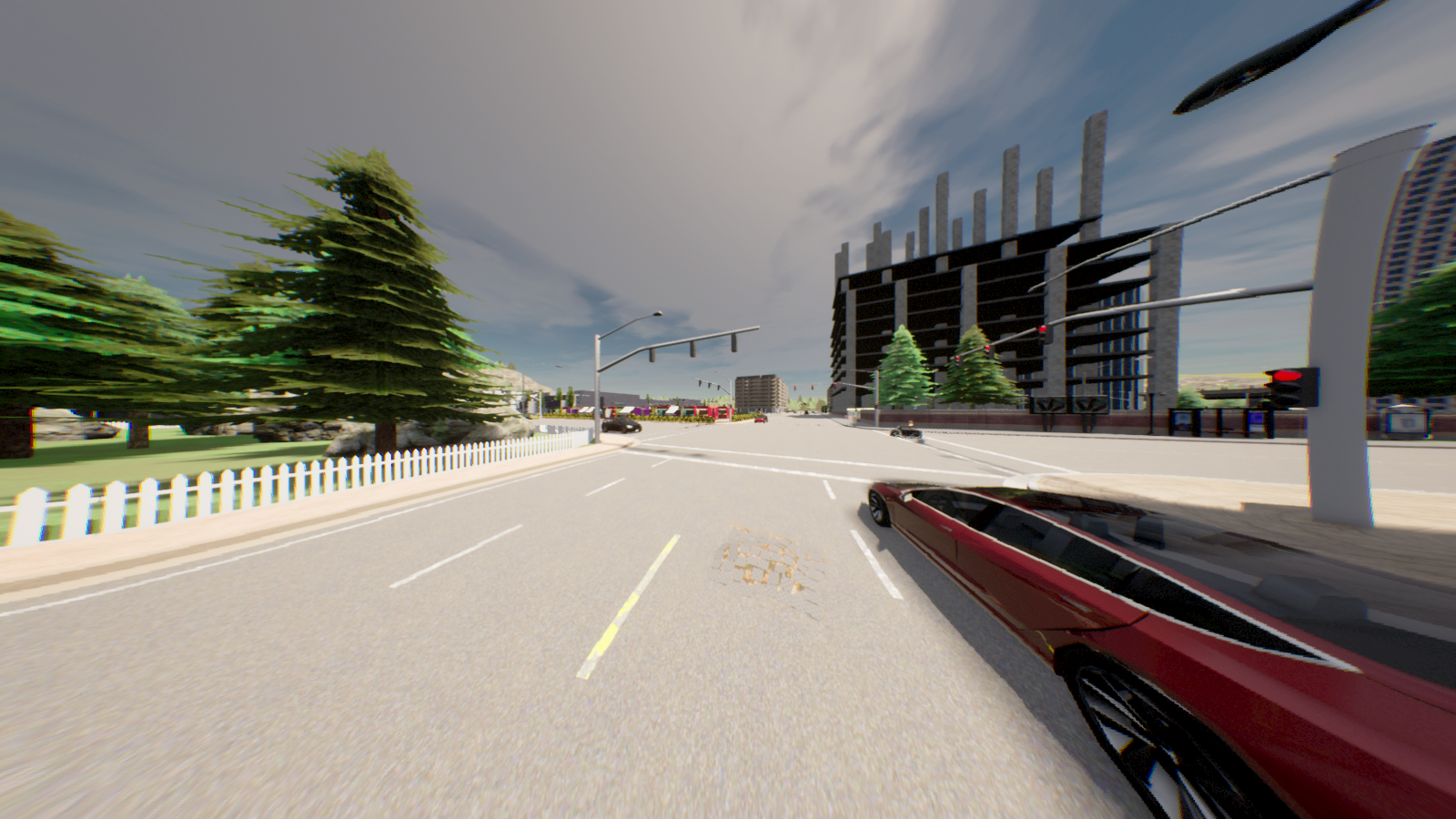} \\
    & Front RGB Camera & RGB Topdown & Structured Visualization & Structured Information \\
  \end{tabular}
  \vspace{3pt}
  \caption{Visualizations of different simulation time steps.The last column shows the visualization of the point cloud and component information.}
  \label{fig:prompt_edit_pixel_case2}
\end{figure}

\subsection{Limiation}
Our current system has only been validated on simulation systems. In the future, we plan to conduct sufficient experiments on real vehicles to further demonstrate the robustness of our method.

\subsection{More Result for Visual Analysis of Responses}

Case 1: As show in Figure \ref{fig:prompt_edit_pixel_case2}--\ref{fig:prompt_edit_weight_case2}, when the vehicle turns right, the responses of routing, command, and target point are the most significant. When there are obstacles in the curve, the responses of obstacles and BEV are more significant, followed by routing. When at an intersection, the responses of traffic lights and stop signs increase significantly.

Case 2: As show in Figure \ref{fig:prompt_edit_pixel_case3}--\ref{fig:prompt_edit_weight_case3}, when approaching a stop line that is not at an intersection, apart from the stop sign itself, the response of speed is more noticeable, indicating that the model pays more attention to the current speed when stopping. Additionally, it can be observed that the response of the map peaks when passing through intersections or curves.

\renewcommand{\addFigs}[1]{\includegraphics[width=1\linewidth]{figs/case2/#1.pdf}}
\begin{figure}
  \centering
  \footnotesize
  \setlength\tabcolsep{0.3mm}
  \renewcommand\arraystretch{0.9}
  \begin{tabular}{ccc}
    \addFigs{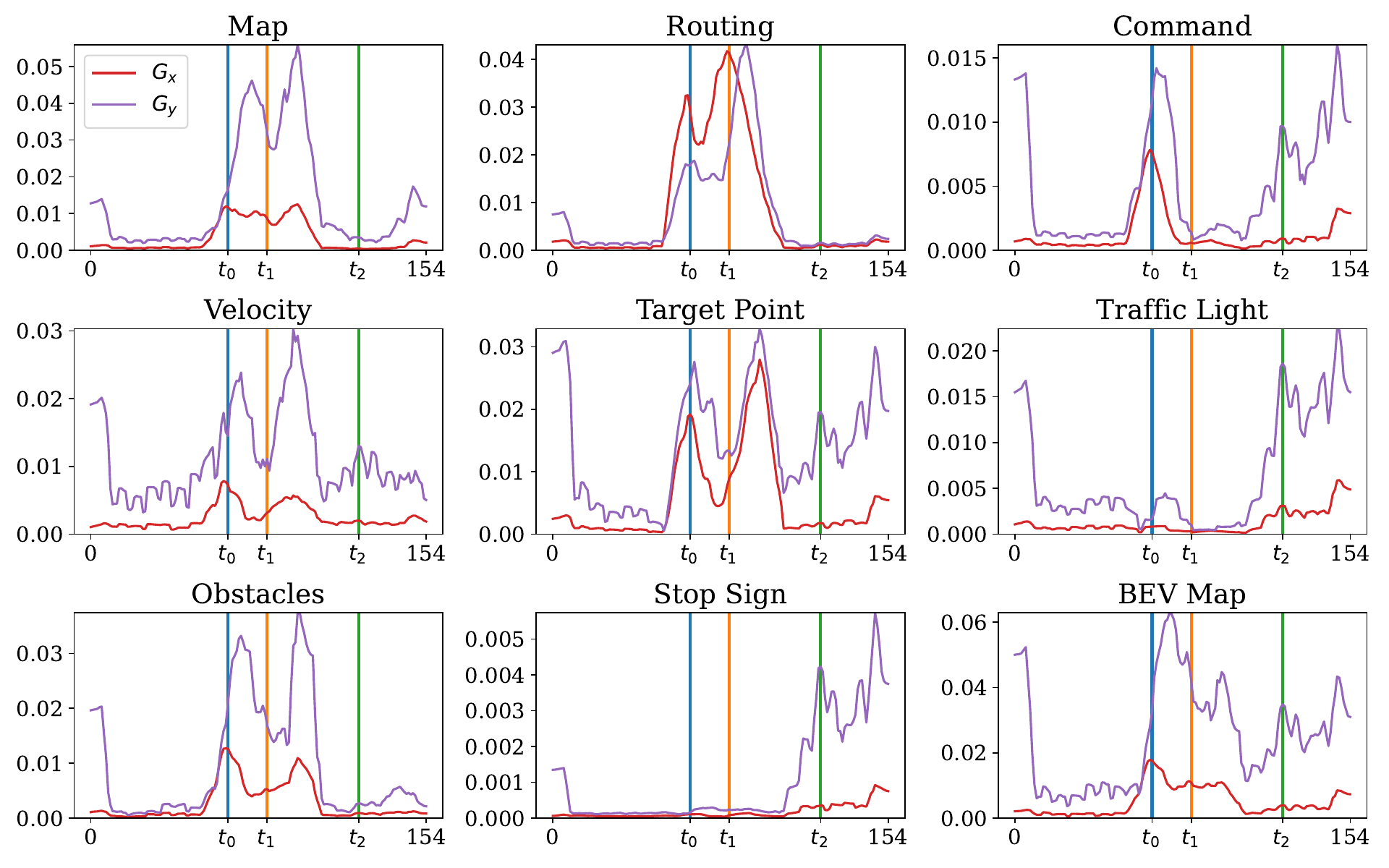}
  \end{tabular}
  \caption{Visualizations of the gradients \emph{w.r.t.} different tokens by simulation time steps. The gradients in the x and y directions are represented by $G_x$ and $G_y$, respectively. The horizontal axis represents the time elapsed along the current route. We sampled three representative moments, denoted as $t_1$, $t_2$, and $t_3$, indicated in the graph by blue, orange, and green vertical lines, respectively. }
  \label{fig:prompt_edit_grad_case2}
\end{figure}

\renewcommand{\addFigs}[1]{\includegraphics[width=1\linewidth]{figs/case2/#1.pdf}}
\begin{figure}
  \centering
  \footnotesize
  \setlength\tabcolsep{0.3mm}
  \renewcommand\arraystretch{0.9}
  \begin{tabular}{cccc}
    \addFigs{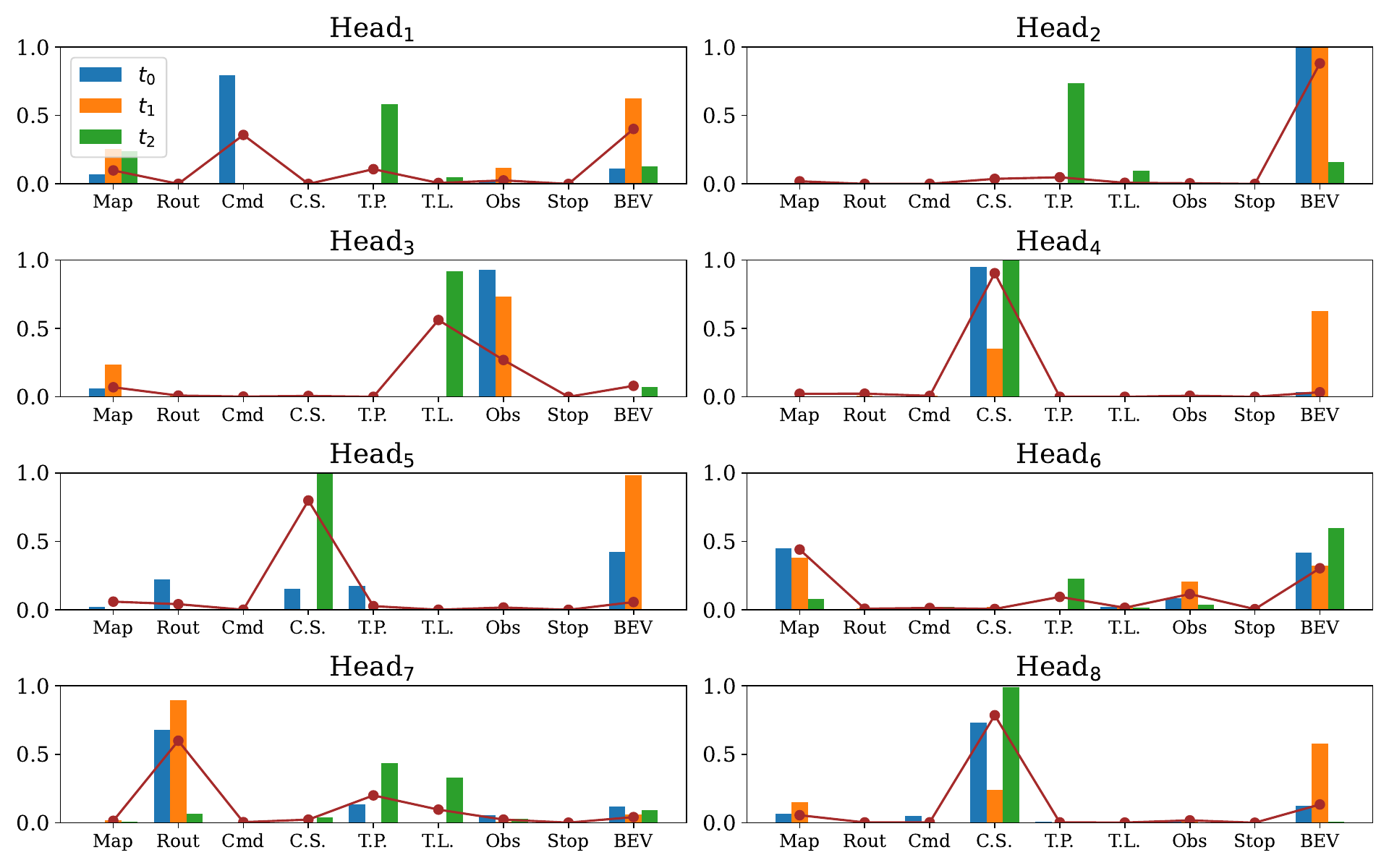}
  \end{tabular}
  \caption{Visualizations of the activation \emph{w.r.t.} different attention heads of different simulation time steps. The three colors in the histogram correspond to the sampling time points in Figure 3. The red line represents the average response value of different components over the observation time interval.}
  \label{fig:prompt_edit_weight_case2}
\end{figure}

\renewcommand{\addFig}[1]{\includegraphics[width=0.245\linewidth]{figs/case3/#1}}
\renewcommand{\addFigCrop}[1]{\includegraphics[trim=350 0 350 0, clip, width=0.245\linewidth]{figs/case3/#1}}
\renewcommand{\addFigs}[1]{\addFigCrop{rgb_front/#1.png} & \addFig{topdown/#1.png} & \addFig{birdview/#1.png} & \addFig{vis/#1.png}}
\begin{figure}
  \centering
  \footnotesize
  \setlength\tabcolsep{0.3mm}
  \renewcommand\arraystretch{1.0}
  \begin{tabular}{ccccc}
    \rotatebox{90}{~~~~~~~~~~~~$t_0=13$} & 
    \addFigs{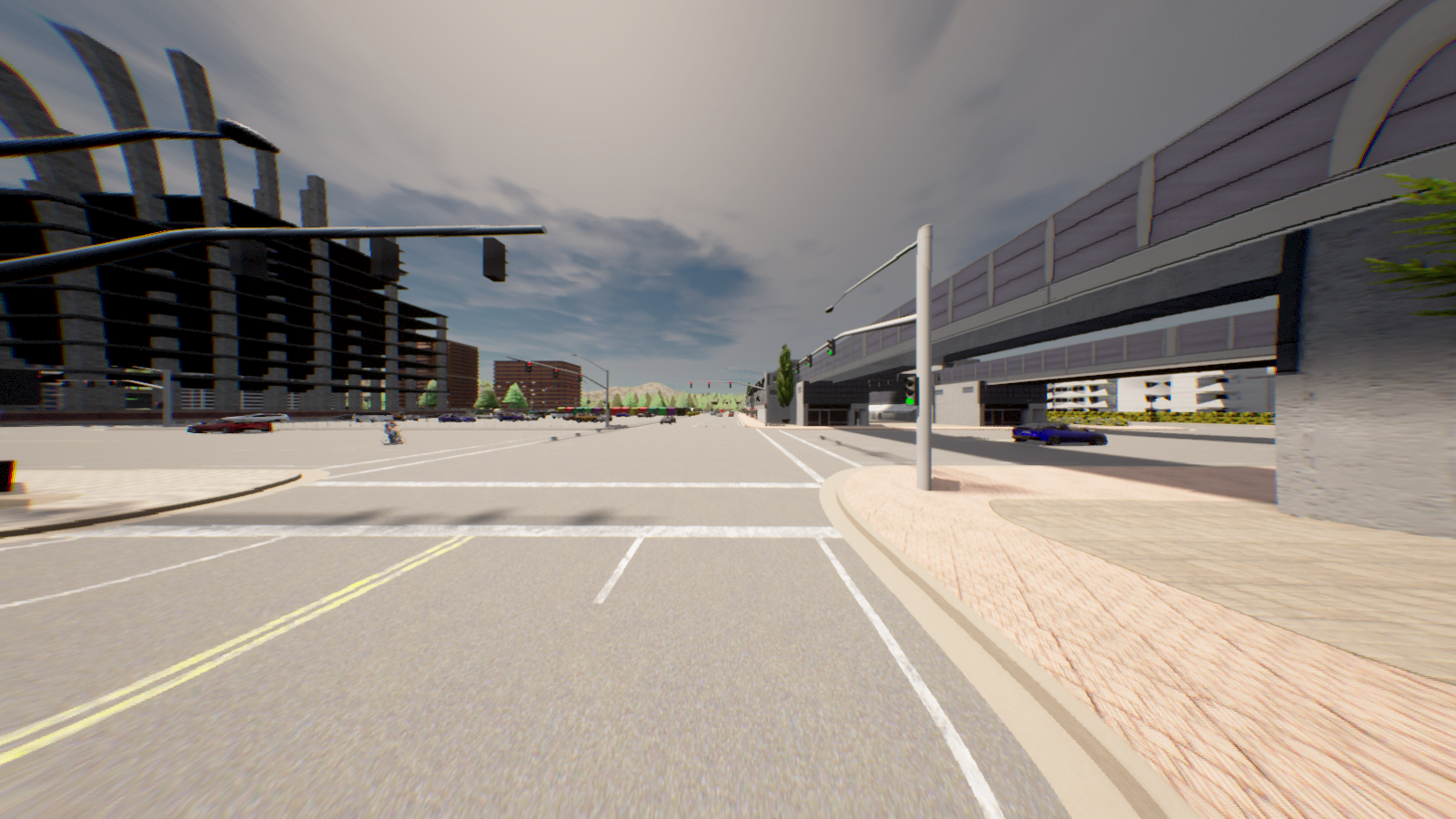} \\
    \rotatebox{90}{~~~~~~~~~~~~$t_1=50$} & 
    \addFigs{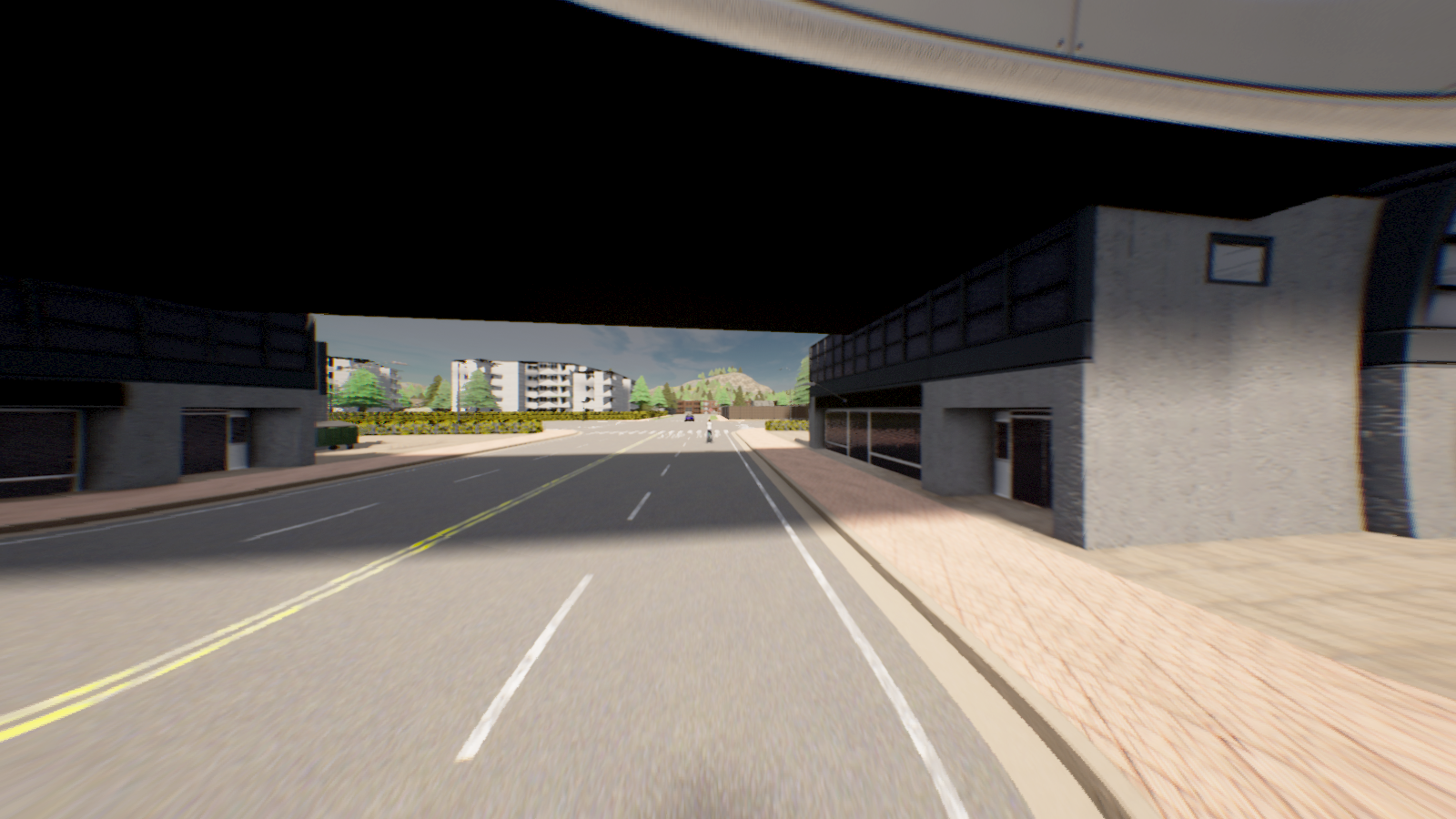}  \\
    \rotatebox{90}{~~~~~~~~~~~~$t_2=107$} & 
    \addFigs{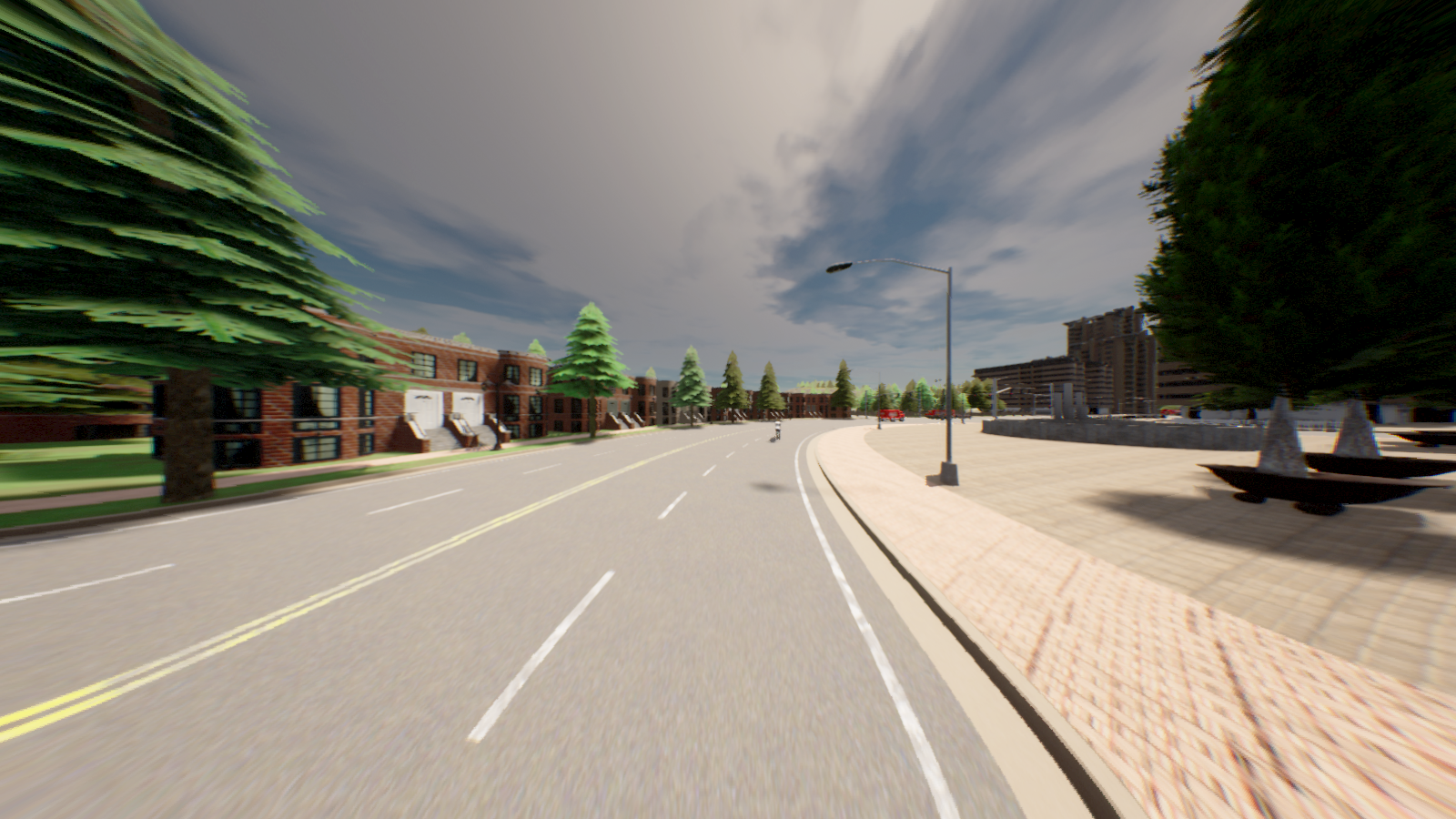} \\
    \rotatebox{90}{~~~~~~~~~~~~$t_2=194$} & 
    \addFigs{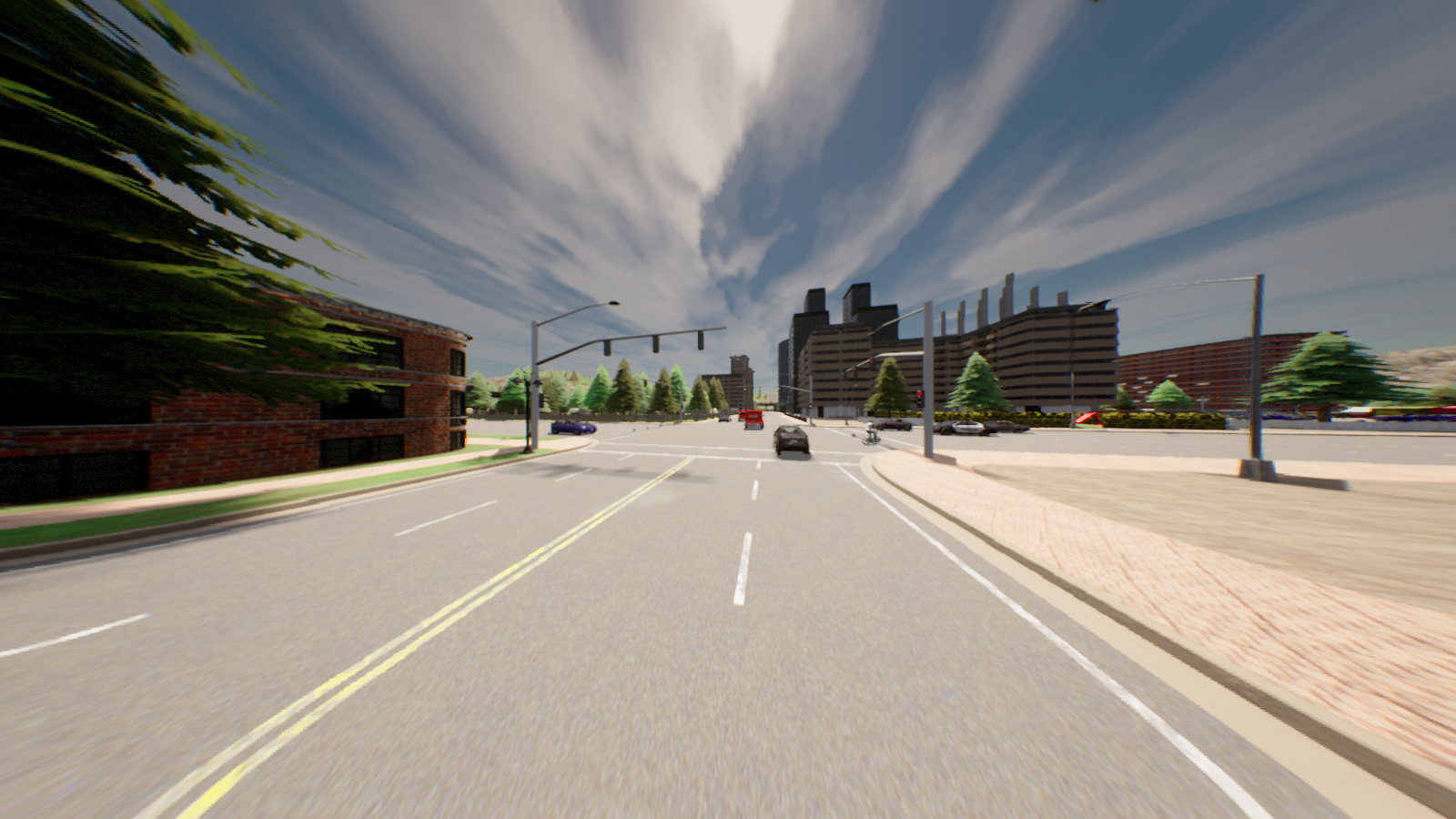} \\
    & Front RGB Camera & RGB Topdown & Structured Visualization & Structured Information \\
  \end{tabular}
  \vspace{3pt}
  \caption{Visualizations of different simulation time steps.The last column shows the visualization of the point cloud and component information. The green curves represent routing, the red dots indicate the target point, the dark blue lines represent the vectorized map, and the light blue rectangles indicate obstacles.}
  \label{fig:prompt_edit_pixel_case3}
\end{figure}

\renewcommand{\addFigs}[1]{\includegraphics[width=1\linewidth]{figs/case3/#1.pdf}}
\begin{figure}
  \centering
  \footnotesize
  \setlength\tabcolsep{0.3mm}
  \renewcommand\arraystretch{0.9}
  \begin{tabular}{c}
    \addFigs{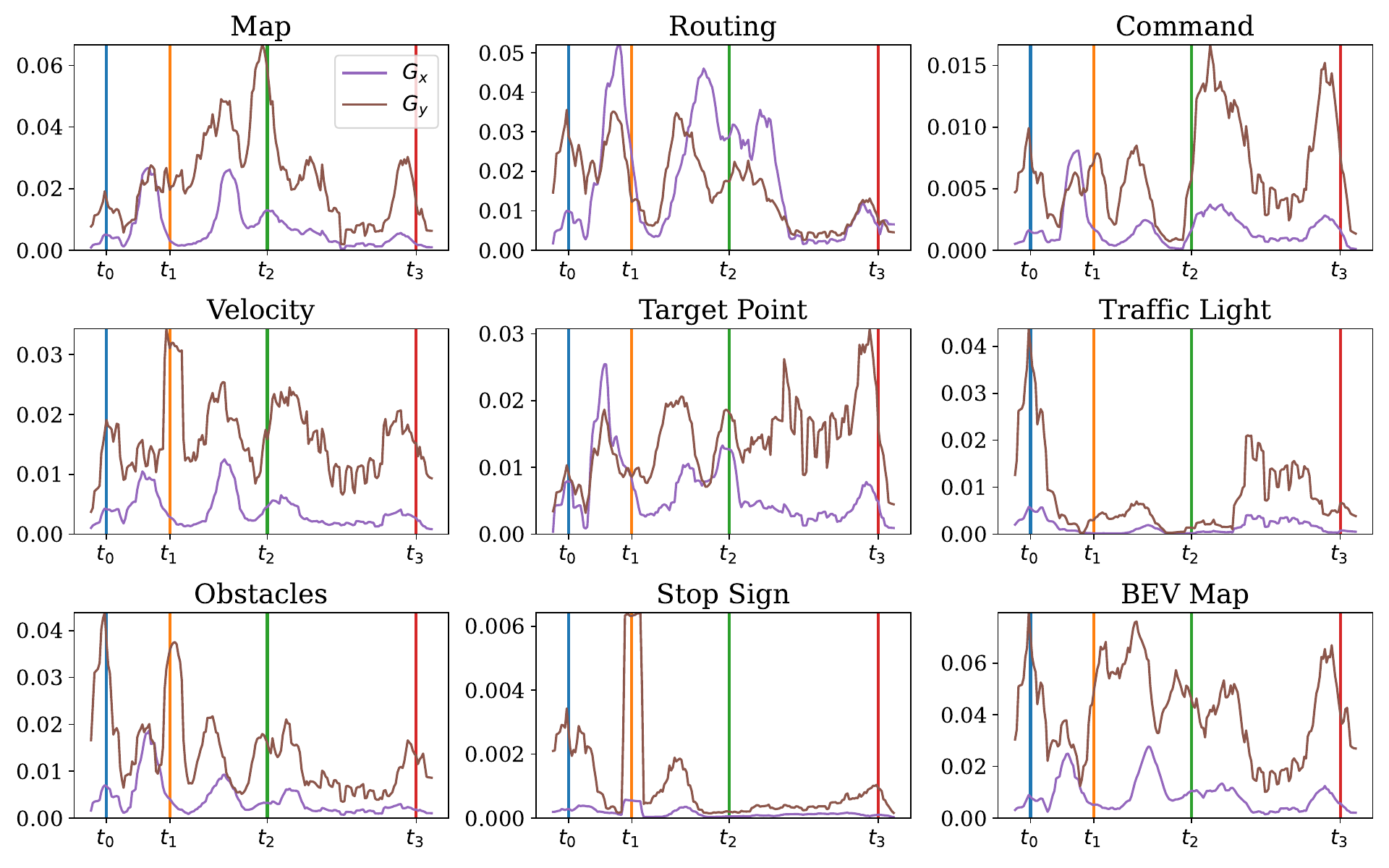}
  \end{tabular}
  \caption{Visualizations of the gradients \emph{w.r.t.} different tokens by simulation time steps. The gradients in the x and y directions are represented by $G_x$ and $G_y$, respectively. The horizontal axis represents the time elapsed along the current route. We sampled three representative moments, denoted as $t_1$, $t_2$, and $t_3$, indicated in the graph by blue, orange, and green vertical lines, respectively. }
  \label{fig:prompt_edit_grad_case3}
\end{figure}

\renewcommand{\addFigs}[1]{\includegraphics[width=1\linewidth]{figs/case3/#1.pdf}}
\begin{figure}
  \centering
  \footnotesize
  \setlength\tabcolsep{0.3mm}
  \renewcommand\arraystretch{0.9}
  \begin{tabular}{c}
    \addFigs{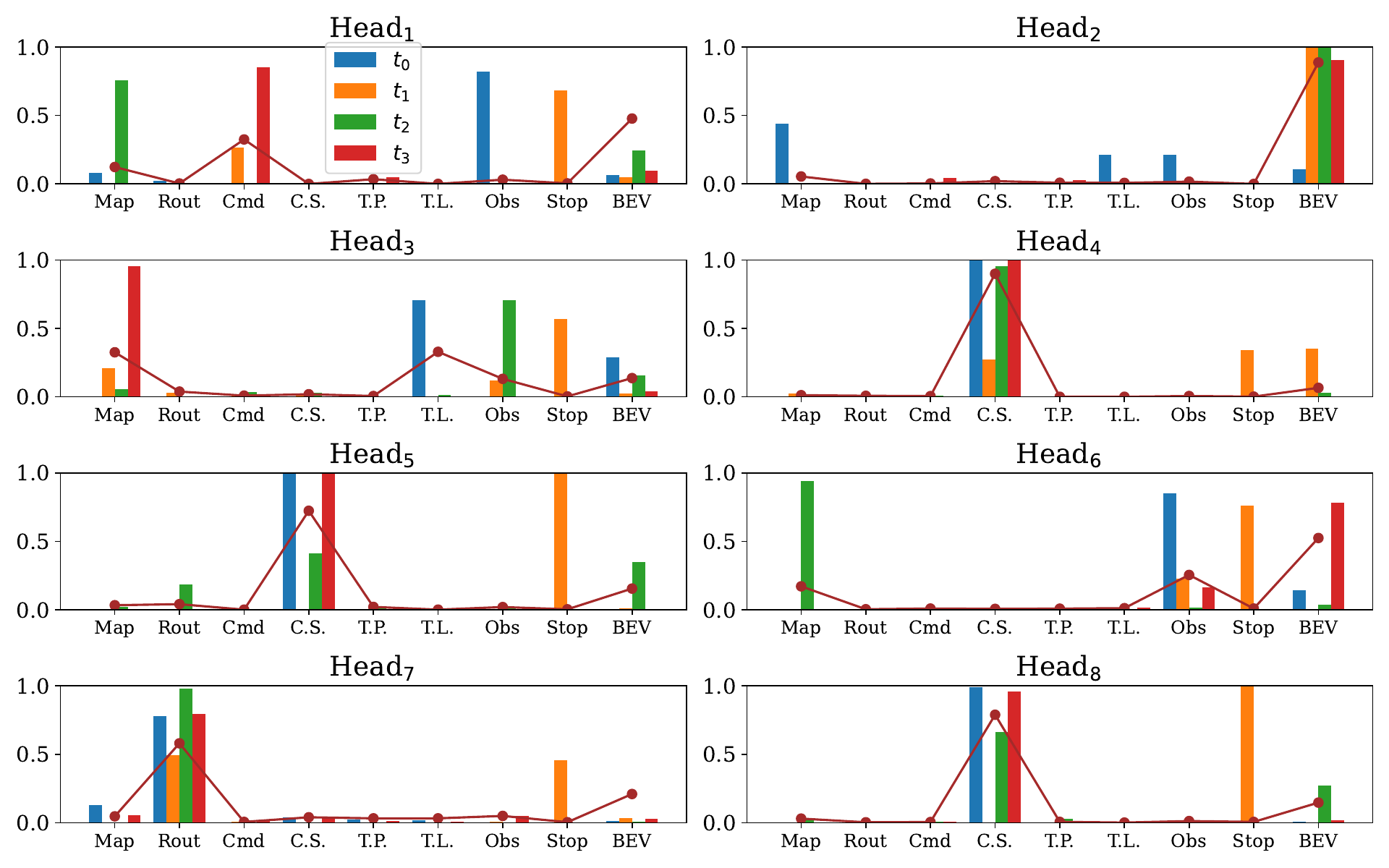}
  \end{tabular}
  \caption{Visualizations of the activation \emph{w.r.t.} different attention heads of different simulation time steps. The three colors in the histogram correspond to the sampling time points in Figure 3. The red line represents the average response value of different components over the observation time interval.}
  \label{fig:prompt_edit_weight_case3}
\end{figure}

\begin{figure}
  \centering
  \footnotesize
  \setlength\tabcolsep{0.3mm}
  \renewcommand\arraystretch{1.0}
  \begin{tabular}{c}
   \includegraphics[width=0.246\linewidth]{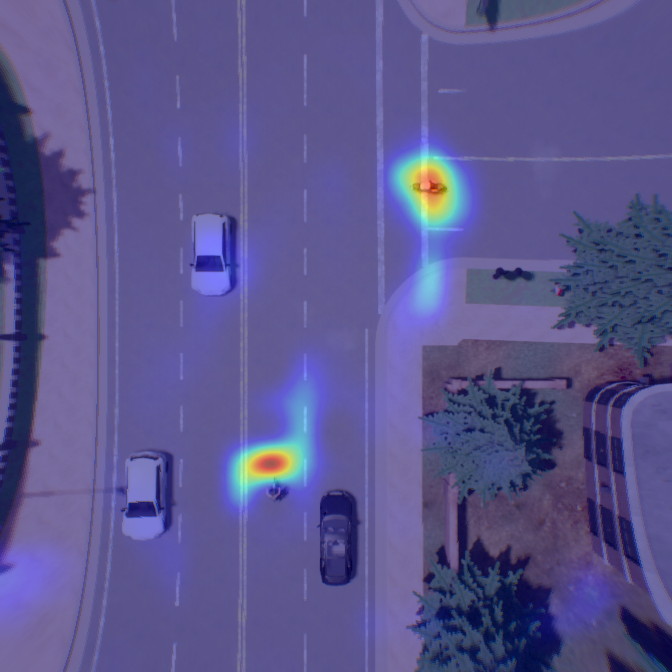}
   \includegraphics[width=0.246\linewidth]{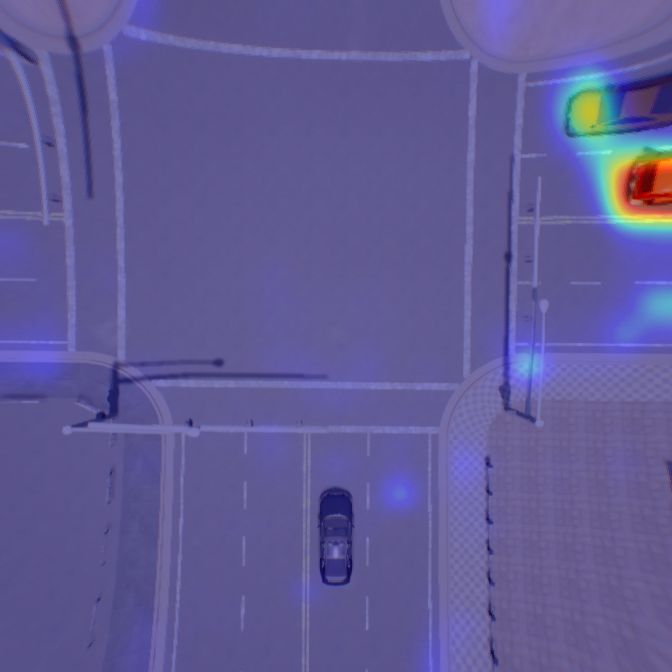}
   \includegraphics[width=0.246\linewidth]{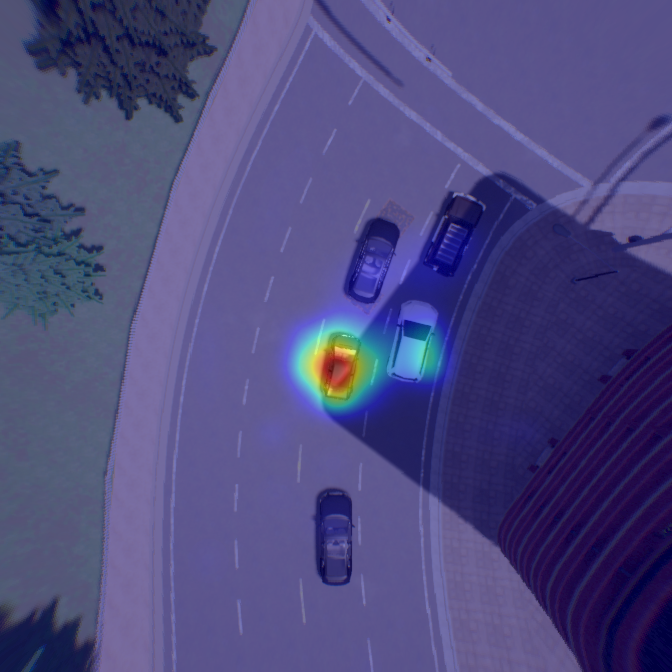}
   \includegraphics[width=0.246\linewidth]{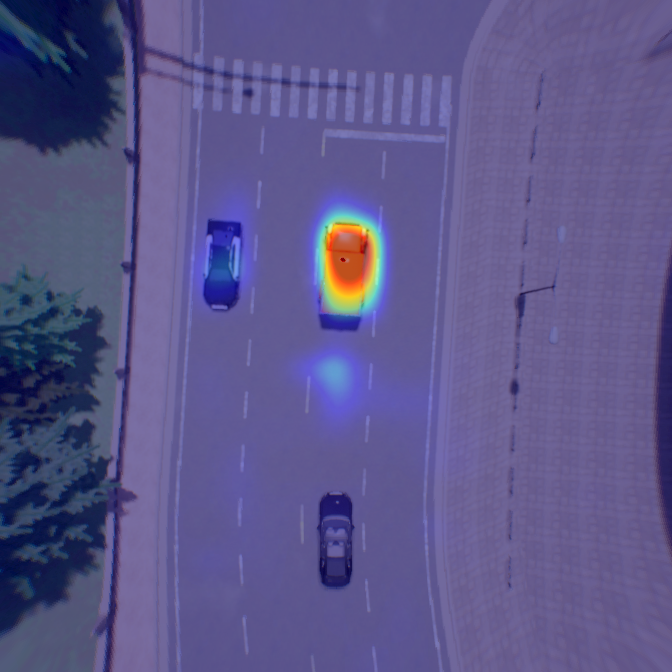} \\
   \includegraphics[width=0.246\linewidth]{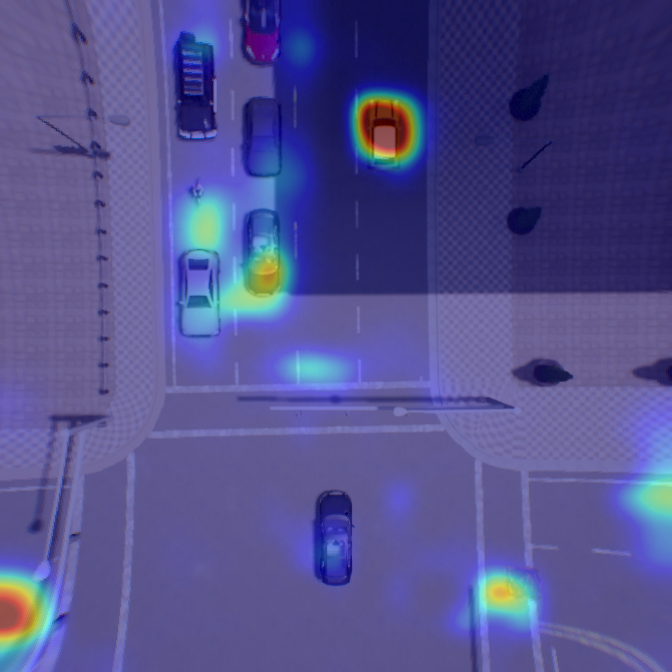}
   \includegraphics[width=0.246\linewidth]{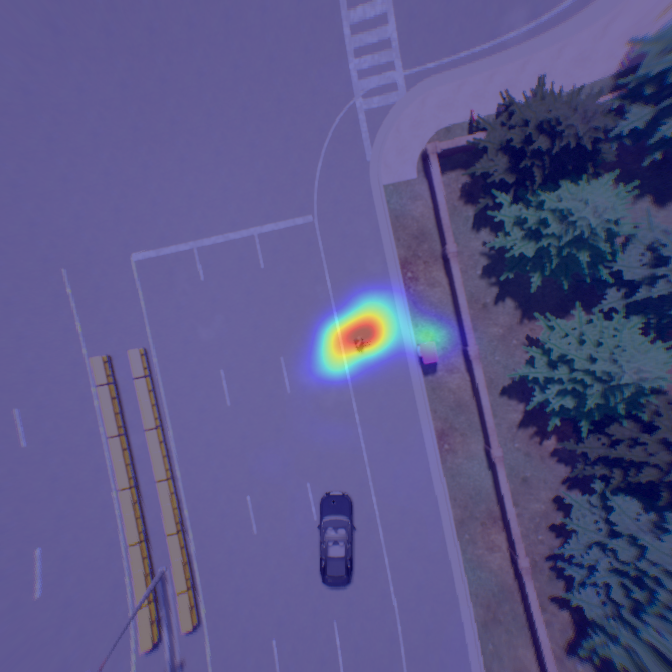}
   \includegraphics[width=0.246\linewidth]{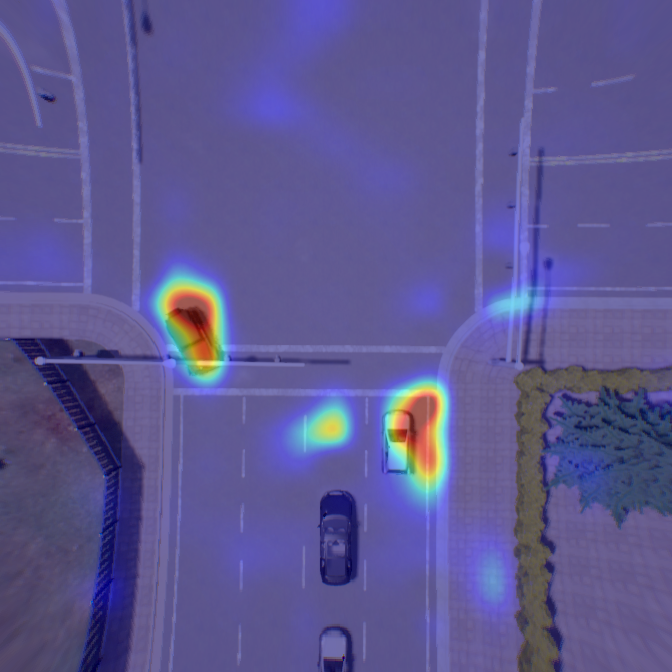}
   \includegraphics[width=0.246\linewidth]{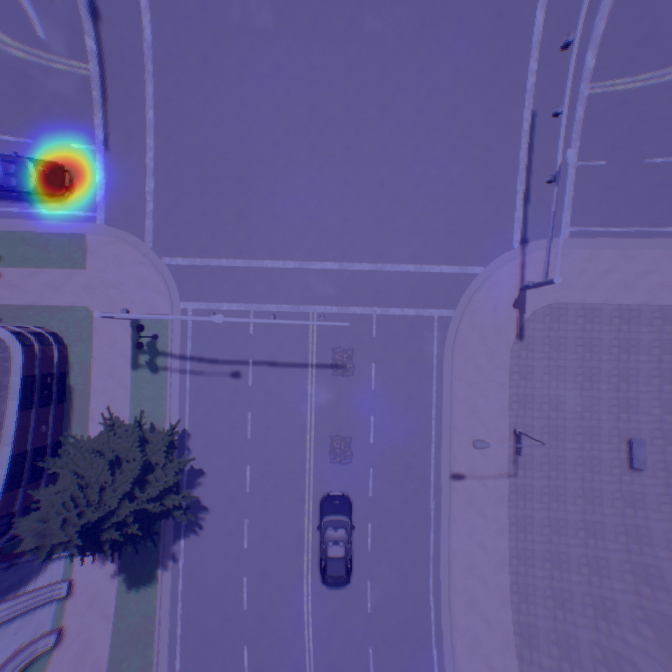} \\
   (a)Activation map of BEV feature.Overlap and aligned with topdown camera in CARLA for visualization \\
   \includegraphics[width=0.33\linewidth]{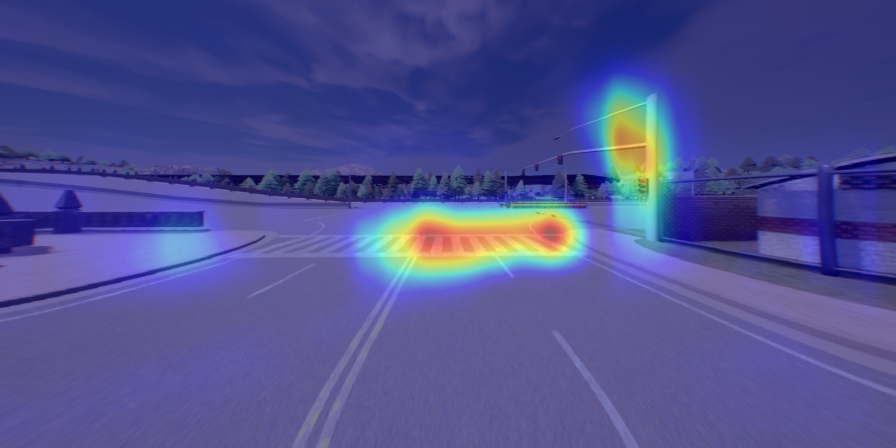}
   \includegraphics[width=0.33\linewidth]{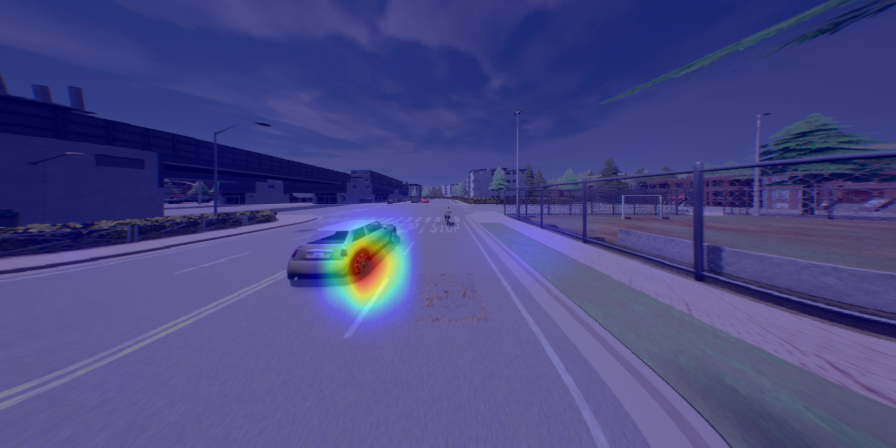}
   \includegraphics[width=0.33\linewidth]{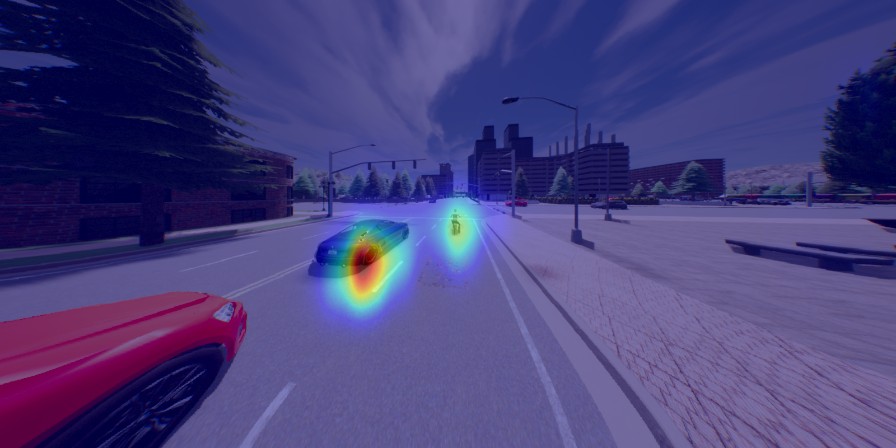} \\
   \includegraphics[width=0.33\linewidth]
   {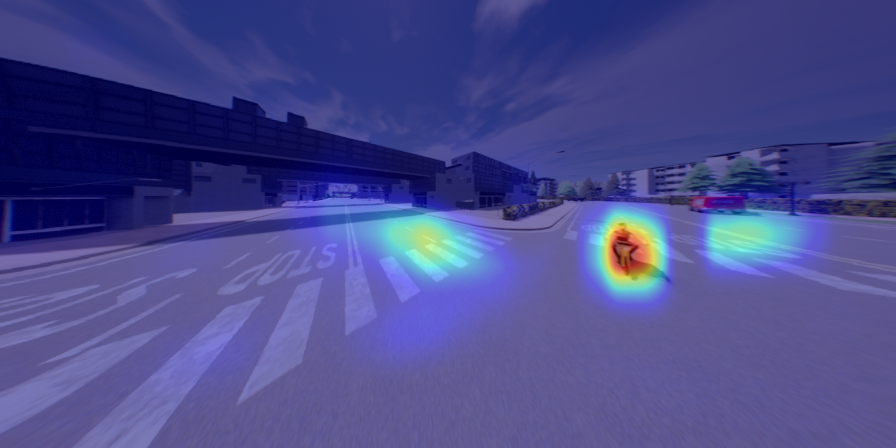}
   \includegraphics[width=0.33\linewidth]{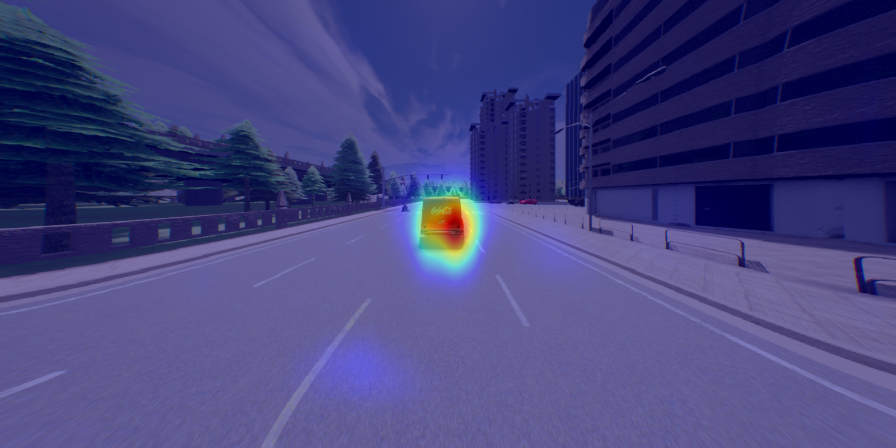}
   \includegraphics[width=0.33\linewidth]{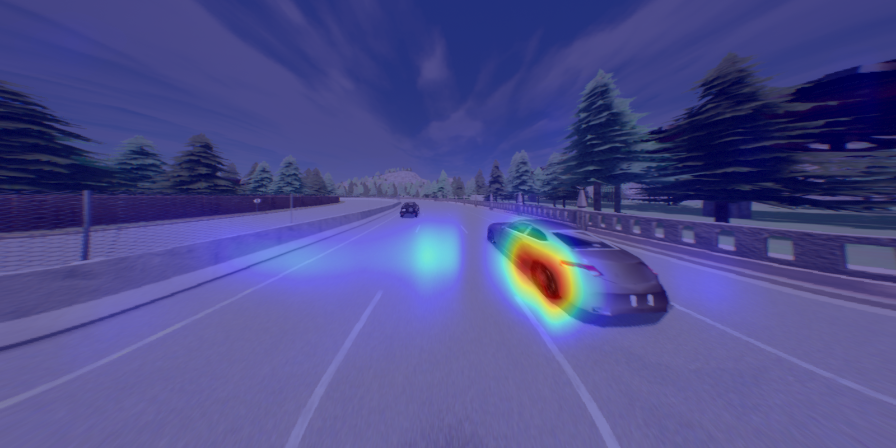} \\
   (b)Activation map of front-view camera feature \\   
  \end{tabular}
  \vspace{3pt}
  \caption{Visualization of activation map}
  \label{fig:activation_map2}
\end{figure}

\subsection{More Result for Activation Map Visualizaion}

In Figure~\ref{fig:activation_map2}, we present additional visualization results of the activation maps for both BEV features and front-view camera features. When obstacles are present, the model demonstrates a significant focus on key obstacles within the current lane and adjacent lanes. Conversely, in the absence of obstacles, the model shifts its attention towards lane markings (e.g., lane center, sidewalks), traffic lights, and signboards.

\end{document}